\pdfoutput=1
\documentclass{article}

\usepackage{microtype}
\usepackage{graphicx}
\usepackage{subcaption}
\usepackage{booktabs}
\usepackage{amsmath}
\usepackage[hyperfootnotes=false]{hyperref}

\usepackage[preprint]{icml2026}
\hypersetup{pdfsubject={Preprint}}

\usepackage{amssymb}
\usepackage{mathtools}
\usepackage{amsthm}
\usepackage{tabularx}
\usepackage{placeins}
\usepackage{algorithm}
\usepackage{algorithmic}
\usepackage[capitalize,noabbrev]{cleveref}

\newcommand{\method}{Mimic-T-Gen}
\newcommand{\data}{Mimic-T-Data}
\newcommand{\adapt}{SONIC-T}

\newcommand{\hstack}[2][c]{\begin{tabular}[b]{@{}#1@{}}#2\end{tabular}}

\newcommand{\cstar}{C^{*}}
\newcommand{\csim}{C^{\mathrm{sim}}}
\newcommand{\chat}{\hat{C}}

\icmltitlerunning{Terrain-as-Mimic: Co-Optimization of Terrain and Policy}

\begin{document}

\twocolumn[
  \icmltitle{Terrain-as-Mimic: Co-Optimization of Terrain and Policy
    for Multi-Contact Physics-Based Humanoid Motion Imitation}

  \icmlsetsymbol{equal}{*}
  \icmlsetsymbol{corresponding}{\textdagger}

  \begin{icmlauthorlist}
    \icmlauthor{Yifei Zhao}{equal,srias,nkl,siist}
    \icmlauthor{Qiwen Xiao}{equal,guohao}
    \icmlauthor{Haozhe Xu}{nkl,control}
    \icmlauthor{Wenhao Feng}{nkl,control}
    \icmlauthor{Chenhui Dong}{srias,nkl,siist}
    \icmlauthor{Zhen Yin}{srias,nkl,control,frontiers}
    \icmlauthor{Zhipeng Wang}{corresponding,srias,nkl,control,frontiers}
    \icmlauthor{Bin He}{corresponding,srias,nkl,control,frontiers}
  \end{icmlauthorlist}

  \icmlaffiliation{srias}{Shanghai Research Institute for Intelligent Autonomous Systems, Tongji University, Shanghai, China}
  \icmlaffiliation{nkl}{National Key Laboratory of Autonomous Intelligent Unmanned Systems, Tongji University, Shanghai, China}
  \icmlaffiliation{siist}{Shanghai Institute of Intelligent Science and Technology, Tongji University, Shanghai, China}
  \icmlaffiliation{guohao}{Guohao College, Tongji University, Shanghai, China}
  \icmlaffiliation{control}{Department of Control Science and Engineering, College of Electronics and Information Engineering, Tongji University, Shanghai, China}
  \icmlaffiliation{frontiers}{Frontiers Science Center for Intelligent Autonomous Systems, Tongji University, Shanghai, China}

  \icmlcorrespondingauthor{Zhipeng Wang\textsuperscript{\textdagger}}{wangzhipeng@tongji.edu.cn}%
  \icmlcorrespondingauthor{Bin He\textsuperscript{\textdagger}}{hebin@tongji.edu.cn}

  \icmlkeywords{terrain-as-mimic, terrain optimization, motion imitation,
    controller--terrain co-optimization, reinforcement learning, human motion}

  \vskip 0.3in
]

\printAffiliationsAndNotice{\icmlEqualContribution}

\begin{abstract}
Learning from human motion data has enabled diverse embodied behaviors on flat
ground, yet motion-only recordings omit the 3D geometry behind stairs, obstacles,
and seats, forcing existing pipelines to filter out such clips or rely on costly
scene capture.
Taking a \emph{terrain-as-mimic} view that sculpts terrain alongside an imitation
policy to match the motion, we introduce \method{} to ground motion-only clips in
physics without scene observations or templates via physical execution feedback and
contact consistency.
Since support surfaces remain imperfect during co-optimization, updating the
imitation policy from physics rollout outcomes lets it learn alternative
foot placements when reference footholds fail.
This adaptation keeps coupled training reliable, letting \method{} scale across
$75.2$\,hours of Nymeria+ and LAFAN1 obstacle motion to yield \data{}, pairing
motion-only clips with sculpted terrains, specialist policies, and rollout-voted
contact labels.
Adapting a pretrained foundation motion tracker into \adapt{} on \data{}
substantially improves terrain-interactive execution under a unified control interface.
\end{abstract}

\section{Introduction}
\label{sec:intro}

Human motion recordings have enabled physics-simulated characters and humanoids to learn
diverse motor skills, from single-clip tracking to generalist policies
\citep{peng2018deepmimic,luo2023phc,luo2024pulse,tessler2024maskedmimic,chen2025gmt,
liao2025beyondmimic,luo2026sonic,ma2026rgmt}.
Yet everyday activities such as climbing stairs, stepping over obstacles, kneeling, and
sitting rely on weight-bearing 3D surfaces.
Optical studios and wearable sensors predominantly yield kinematics-only motion archives
without paired support geometry, leaving terrain-paired data scarce and leading pipelines
to filter out clips with missing supports or floating artifacts
\citep{luo2023phc,tessler2024maskedmimic,guan2026limmt,luo2026sonic,liao2026supermimic}.

Prior terrain-aware pipelines obtain geometry from video reconstruction
\citep{allshire2025videomimic}, paired 3D scans \citep{brandes2026egohtr}, known-scene
augmentation or synthesis \citep{yang2025omniretarget,xie2026grail}, procedural terrains
with kinematically adapted clips \citep{miao2026nexus} or motion priors
\citep{guo2026tgmp,lee2026echo,zhu2026trip}, manual obstacle alignment refined in
simulation \citep{chen2026lightparkour}, or static kinematic contact fitting before policy
training \citep{chen2026scenebot,cheng2026gigabrain}.
Motion-only tracks lack external scene observations, while kinematics alone leaves support
ambiguous because a low swing limb resembles a loaded stance and retargeting offsets,
wearable drift, and contact compliance shift effective support height.
Motion-conditioned 3D scene synthesis instead targets semantically plausible indoor
objects, predicting oriented object boxes \citep{nie2022pose2room}, placing or generating
furniture layouts from human contacts and free-space motion
\citep{ye2022summon,yi2023mime}, or optimizing library mesh placement jointly with a
controller \citep{li2024physics}.
Acquiring terrain-interactive humanoid control, however, depends on physically calibrated
support elevation and unobstructed swing paths rather than object-level semantics.

\begin{figure*}[t]
  \centering
  \includegraphics[width=0.93\linewidth]{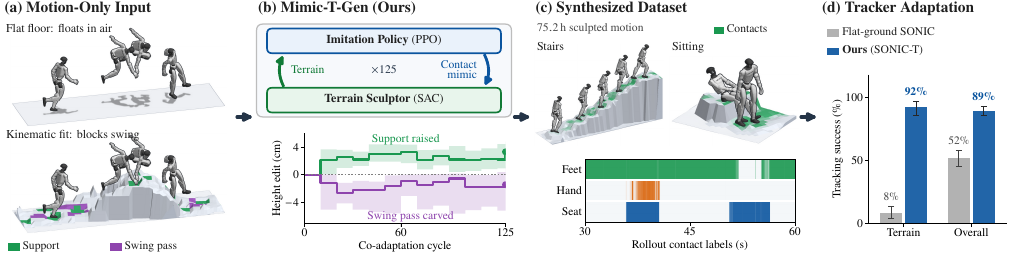}
  \caption{End-to-end overview. (a)~Poses float on a flat floor, while a kinematic fit
    blocks the swing pass. (b)~\method{} co-optimizes PPO imitation and SAC sculpting via
    contact-mimic rewards to raise supports and carve passes. (c)~\data{} ($75.2$\,h across
    Nymeria+ and LAFAN1) pairs sculpted terrains with rollout contact labels, enabling
    (d)~\adapt{} ($8.1\%\!\to\!92.2\%$ terrain success).}
  \label{fig:teaser}
\end{figure*}

Rather than imitating motion on a fixed scene, we take a \emph{terrain-as-mimic} view in
which the support surface itself is optimized alongside the policy to match the contacts
and clearances implied by the reference motion.
We instantiate this view in \method{}, \textbf{Mimic} \textbf{T}errain
\textbf{Gen}eration, co-optimizing a dense support heightfield and an imitation policy
from motion-only trajectories without scene observations or templates (\cref{fig:teaser}).
Initialized at the geometric ceiling, the lower envelope of the motion's swept collision
volume, \method{} couples an on-policy PPO imitation policy with an off-policy SAC terrain
sculptor guided by contact-mimic rewards over a differentiable brush, supplemented with
analytic shape regularization to clear non-supporting passes and calibrate supports.

Sustaining this coupled loop requires an inner learner that stays viable on imperfect,
evolving surfaces.
PPO \citep{schulman2017ppo} samples joint commands and estimates their advantages from
physics rollouts, learning alternative foot placements during training and executing
without online search.
Obstacle co-optimization and stair perturbations (\cref{fig:stairs-perturb},
\cref{app:exp-planner}) show PPO learning alternative footsteps where our world-model
tracker inspired by \citet{xu2022shac} and \citet{liao2026supermimic} fails.

Scaling \method{} across $75.2$\,h of motion tracks alone from Nymeria+
\citep{detone2026nymeriaplus}, an extension of Nymeria \citep{ma2024nymeria}, and LAFAN1
obstacle clips \citep{harvey2020lafan1} yields \data{}, pairing clips with sculpted
terrains, specialists, and rollout-voted contact labels.
Beyond per-clip specialists, we adapt frozen SONIC \citep{luo2026sonic} into \adapt{},
a contact-conditioned low-rank adapter and predictor for unified terrain control.

Our contributions are as follows.
\begin{itemize}
  \item We formulate physical grounding of motion-only clips as \emph{terrain-as-mimic}
        co-optimization of a heightfield and policy (\cref{sec:problem}), coupling PPO
        tracking with off-policy SAC sculpting in \method{} (\cref{sec:method}).
  \item We analyze learning at both timescales, showing how PPO adapts foot
        placement on imperfect support (\cref{sec:planning,sec:exp-planner})
        and how decoupled priors and replay speed sculpting (\cref{sec:exp-sculptor}).
  \item We scale \method{} across $75.2$\,h of Nymeria+ and LAFAN1 obstacle
        motion to construct \data{} with rollout-voted contact labels and adapt pretrained
        SONIC into \adapt{} (\cref{sec:data}).
\end{itemize}

\section{Dual-Policy Formulation}
\label{sec:problem}

\subsection{Two-Timescale Policy Co-Optimization}
Given a motion-only reference clip $m$ of length~$T$ and a target robot embodiment
without scene observations, terrain templates, or object libraries, our
\emph{terrain-as-mimic} formulation jointly optimizes a dense heightfield
$h\in\mathbb{R}^{N\times N}$ and a motion-imitation policy so that the robot physically
reproduces $m$ in simulation. We cast this task as a coupled two-timescale learning
problem between two policies:
\begin{equation}\label{eq:two-policies}
  \begin{aligned}
    a_t &\sim \pi_\theta(\cdot\mid o_t), & s_{t+1} &= F\bigl(s_t,a_t;\mathcal{E}(h_k,u_k)\bigr),\\
    u_k &\sim \pi_\phi(\cdot\mid s^{\mathrm{ter}}_k), & h_{k+1} &= \mathcal{E}(h_k,\bar u_k).
  \end{aligned}
\end{equation}%
\begin{figure*}[t]
  \centering
  \includegraphics[width=0.90\linewidth]{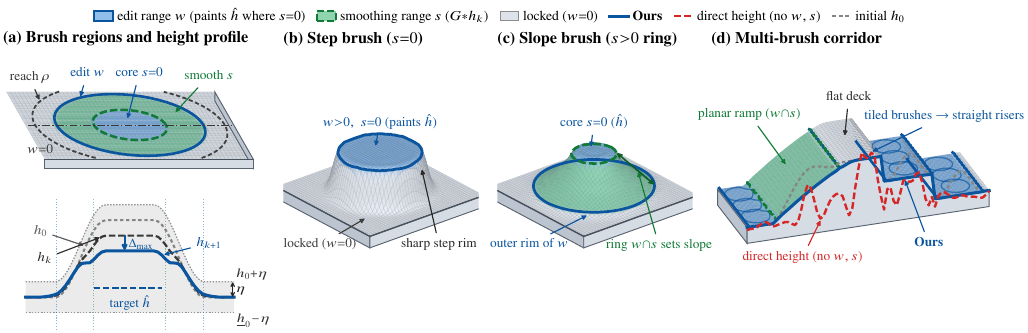}
  \caption{Differentiable two-range terrain edit operator in \cref{eq:edit}, computed on a
    $0.08$\,m grid ($0.12$\,m control lattice and Gaussian blur standard deviation,
    $\eta{=}\Delta_{\max}{=}5$\,cm).
    (a)~Regions of one circular brush (top) and heights along its dash-dotted diameter
    (bottom), labelled with the notation of \cref{eq:edit}.
    (b,~c)~Single circular brush with $s{=}0$ and with a concentric smoothing ring $s{>}0$
    inside~$w$.
    (d)~Multi-brush forward corridor comparing \textbf{Ours} against direct height edits
    without $(w,s)$ under a $125$-cycle budget.}
  \label{fig:edit}
\end{figure*}%
Here $\pi_\theta(a_t\mid o_t)$ is a fast, step-level \emph{motion-imitation policy}
tracking $m$'s kinematics at $50$\,Hz under rigid-body contact dynamics~$F$,
$\pi_\phi(u_k\mid s^{\mathrm{ter}}_k)$ is a slow \emph{terrain-sculpting policy} adapting
the surface across \emph{co-adaptation cycles} to match $m$'s implied supports, and
$\mathcal{E}$ is a differentiable terrain edit operator. Within each cycle~$k$, the sampled edit $\mathcal{E}(h_k,u_k)$ is held fixed
across $I{=}96$ inner imitation updates so $\pi_\theta$ adapts its gait while
aggregating survival and spatial contact evidence, after which the terrain
keeps only the mean-action edit $h_{k+1}=\mathcal{E}(h_k,\bar u_k)$ so exploration noise
never accumulates. A $12{,}000$-iteration run yields millions of control steps~$t$ for
$\pi_\theta$ yet only about $125$ cycles~$k$ per terrain, with episode-level execution
non-differentiable in~$h$. We therefore pair the two policies with timescale-matched
critics (\cref{sec:population}), where an on-policy step critic $V(s_t)$ trains
$\pi_\theta$ via PPO \citep{schulman2017ppo} with generalized advantage estimation
\citep{schulman2016gae} and an off-policy cycle critic $Q(s^{\mathrm{ter}}_k,u_k)$ trains
$\pi_\phi$ via SAC \citep{haarnoja2018sac} from replay-buffered contact-mimic and tracking
rewards, with analytic shape regularization on the actor via~$\mathcal{E}$.

\subsection{Motion-Imitation Policy $\pi_\theta$ (PPO)}
\label{sec:tracker}
At the $50$\,Hz control step, $\pi_\theta(a_t\mid o_t)$ builds directly on BeyondMimic
\citep{liao2025beyondmimic} for the $29$-DoF Unitree G1. A three-layer MLP actor
$\pi_\theta$ maps proprioception and time-aligned reference targets to PD joint setpoints,
paired with an MLP value critic $V(s_t)$, across $8{,}192$ parallel environments over $10$-s
episodes with adaptive reference-state initialization, terminating when torso or
end-effector height deviates from the reference by more than $\sigma{=}0.25$\,m, torso
orientation deviates beyond a fixed bound, or simulation state diverges.
We make three multi-contact adaptations.
\begin{itemize}
  \item \emph{Terrain perception.} Both $\pi_\theta$ and $V(s_t)$ observe an egocentric
        $17\times11$ height scan over $1.6\times1.0$\,m, following perceptive legged and
        character control \citep{rudin2022learning,tessler2024maskedmimic}
        so one shared $\pi_\theta$ adapts across co-evolving candidate terrains.
  \item \emph{Drift-tolerant pose alignment.} Body-pose rewards vertically align the
        reference's lowest body to the robot's at each step to tolerate inertial-capture
        drift, while torso height stays absolute to bind the robot to~$h$.
  \item \emph{Multi-body contact.} Undesired-contact penalties are disabled to allow
        sitting, kneeling, and hand support.
\end{itemize}

\subsection{Terrain-Sculpting Policy $\pi_\phi$ (SAC)}
\label{sec:editor}
Starting from $h_0\in\mathbb{R}^{N\times N}$ (\cref{sec:ceiling}),
$\pi_\phi(u_k\mid s^{\mathrm{ter}}_k)$ updates $h_k$ each cycle~$k$ via off-policy
SAC with clipped double Q-learning \citep{fujimoto2018td3,haarnoja2018sac,haarnoja2018sacapps},
guided by contact-mimic and tracking rewards plus analytic priors on
$\mathcal{E}(h_k,\bar u_k)$ (\cref{sec:objective,sec:population}).%
\begin{figure*}[t]
  \centering
  \includegraphics[width=0.90\linewidth]{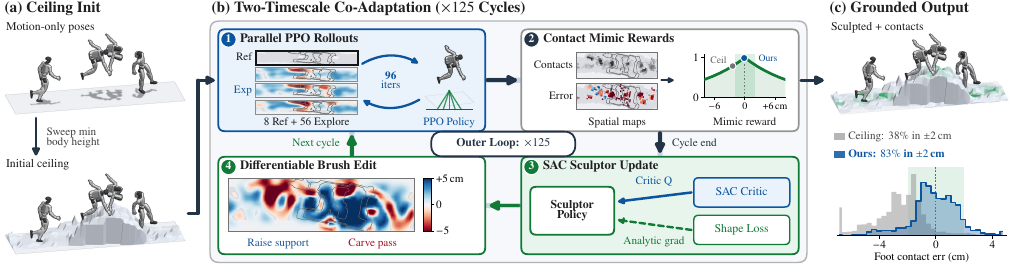}
  \caption{Workflow of \method{} on a $6.4$-s speed vault. (a)~Poses sweep initial
    ceiling $h_0$ (\cref{sec:ceiling}). (b)~Outer loop ($\times 125$ cycles,
    \cref{sec:population}): (1)~$96$-iteration PPO rollouts on fixed tiles, (2)~contact
    maps and contact-mimic rewards, (3)~SAC update with shape priors
    (\cref{sec:objective}), and (4)~brush edits. (c)~Grounded output.}
  \label{fig:overview}
\end{figure*}

\paragraph{State $s^{\mathrm{ter}}_k$.}
We discretize $h_k$ on a $10\times10$\,m grid at $0.08$\,m spacing ($N{=}126$). To show
$\pi_\phi$ cycle-$k$ geometry and support outcomes, each robot--terrain contact maps from
its 3D point to its grid cell, stacking five $\mathbb{R}^{N\times N}$ maps into
\begin{equation}\label{eq:ter-state}
  s^{\mathrm{ter}}_k = \bigl[\,\tilde h_k,\; d_k,\; \tilde n_k,\; A_k,\; e^{+}_k\,\bigr]
  \in \mathbb{R}^{5\times N\times N}.
\end{equation}
\begin{itemize}
  \item \emph{Current height and reference offset} $\tilde h_k, d_k$. The surface height
        $h_k$ linearly normalized to $\tilde h_k\in[0,1]^{N\times N}$ over a fixed
        vertical interval $[z_{\min},z_{\max}]$ spanning the reference motion's lowest to
        highest body positions with collision margins, paired with the scaled displacement
        $d_k=(h_k-h_0)/\sigma$ from the initial reference surface~$h_0$.
  \item \emph{Contact density} $\tilde n_k$. Peak-normalized cell contact count
        $\tilde n_k=n_k/\max(n_k)$, indicating which cells the robot actually stepped or
        rested on during cycle~$k$ while leaving untouched aerial cells at zero.
  \item \emph{Per-cell contact-mimic advantage} $A_k$. For cells visited at least eight
        times during cycle~$k$, we average how closely their contacts match the reference
        contact height using the contact-mimic reward from \cref{sec:objective}, then
        standardize those cell averages to zero mean and unit variance across the visited
        cells, clipped to $[-3,3]$. Sparsely visited cells remain zero, while positive and
        negative values distinguish well-calibrated support cells from overly high or low ones.
  \item \emph{First-step vertical lift} $e^{+}_k$. A reset places the robot exactly on the
        reference pose, so a body that ends up above its reference after the first
        simulation step has been lifted by the terrain rather than moved by the policy.
        Recording this lift at contacted cells marks terrain standing above the motion.
\end{itemize}

\paragraph{Action $u_k$ and differentiable edit operator $\mathcal{E}(h_k,u_k)$.}
Predicting independent cell heights directly leaves bumpy pillars and trenches within about
$125$ sculpting cycles (\cref{fig:edit}d), since uncoordinated height targets neither
interpolate planar slopes across unvisited gaps nor freeze validated step edges. Instead, a
convolutional actor samples a $\tanh$-squashed Gaussian action
$u_k\in\mathbb{R}^{3\times84\times84}$, whose squashed mean is $\bar u_k$, on a $0.12$\,m
control lattice, following
\citet{haarnoja2018sac,haarnoja2018sacapps}, who use invertible $\tanh$ squashing to
enforce finite continuous action bounds while preserving reparameterized policy gradients
and closed-form log-densities. Bicubic upsampling onto the $N\times N$ terrain grid
followed by smooth channel mappings keeps $\mathcal{E}$ end-to-end
differentiable and decodes any lattice action---$u_k$ for the simulated exploratory
surface $\mathcal{E}(h_k,u_k)$ and $\bar u_k$ for the retained update
$h_{k+1}=\mathcal{E}(h_k,\bar u_k)$---into the three spatial brush fields in
\cref{fig:edit}a, namely an absolute target height
$\hat h\in[\underline{h}_0-\eta,\,h_0+\eta]$, an edit range $w\in[0,\rho]$, and a smoothing
range $s\in[0,1]$. Here $\underline{h}_0=\min(h_0,h_{\mathrm{floor}})$ preserves motions
dipping below the floor, $\eta{=}5$\,cm allows upward support calibration while letting
targets reach the floor beneath metre-tall ceiling walls, and the reach mask
$\rho\in\{0,1\}^{N\times N}$ dilates cycle~$k$'s swept footprint by $0.24$\,m so untouched
regions remain locked at $w{=}0$.

\begin{subequations}\label{eq:edit}
Inside the edit range $w$, each cell blends the target height $\hat h$ with the
Gaussian-smoothed surface $G\ast h_k$ of standard deviation $0.12$\,m according to $s$, and
filtering the masked residual with~$G$ gives the proposed height change
\begin{equation}\label{eq:edit-delta}
  \Delta h_k = G \ast \Bigl(w\bigl[(1 - s)\,\hat h + s\,(G \ast h_k) - h_k\bigr]\Bigr).
\end{equation}
Since $\hat h$ can lie more than a metre below $h_k$, applying $\Delta h_k$ in one cycle
could drop a load-bearing tread to the floor and erase the contacts needed to restore it.
$\mathcal{E}$ therefore limits each cycle's update to $\Delta_{\max}{=}5$\,cm via a smooth
$\tanh$ step and clips spillover to the height bounds,
\begin{equation}\label{eq:edit-step}
  h_{k+1} = \operatorname{clip}_{[\underline{h}_0-\eta,\,h_0+\eta]}
    \!\Bigl(h_k + \Delta_{\max}\tanh\!\Bigl(\tfrac{\Delta h_k}{\Delta_{\max}}\Bigr)\Bigr),
\end{equation}
where $\tanh$ keeps $\partial h_{k+1}/\partial\Delta h_k>0$ so shape losses
backpropagate through $h_{k+1}=\mathcal{E}(h_k,\bar u_k)$ into $\pi_\phi$.
\end{subequations}

Filtering the masked residual in \cref{eq:edit-delta} rather than $h_k$ itself ensures
that $w{=}0$ gives $\Delta h_k{=}0$ and yields the exact identity update $h_{k+1}{=}h_k$,
preventing validated stairs from diffusing across cycles. Within $w{>}0$, the gap between
$w$ and $s$ controls transition geometry. Setting $s{=}0$ drives cells onto $\hat h$
against a locked $w{=}0$ rim to form sharp steps (\cref{fig:edit}b), whereas a concentric
$s{>}0$ ring relaxes $w\cap s$ via $G\ast h_k$ into a planar slope whose ring width sets
its grade (\cref{fig:edit}c). Tiling circular brushes merges curved footprints into
straight ramps, decks, and risers (\cref{fig:edit}d).

\section{\method{} Design and Co-Optimization}
\label{sec:method}

\Cref{fig:overview} summarizes the coupled optimization workflow.

\subsection{Geometric Ceiling Initialization}
\label{sec:ceiling}
\begin{figure}[t]
  \centering
  \includegraphics[width=0.66\linewidth]{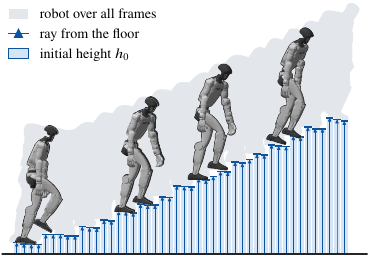}
  \caption{Initial terrain $h_0$ under the first flight of a stair ascent (frames 0--560,
    $11.2$\,s, side view). One ray per $0.08$\,m column rises from the floor and stops at the
    lowest point of the robot over all frames, recovering $15$ treads $\sim\!9$\,cm apart.}
  \label{fig:ceiling}
\end{figure}

Discovering metre-tall stairs or seats by growing terrain upward from a flat floor fails
because the robot falls in mid-air before encountering elevated support. We therefore start
every grid cell at the \emph{geometric ceiling}. A ray rises from the floor through the cell
and stops at the lowest point the robot's collision geometry reaches above the cell over
all reference frames, setting the cell's initial height~$h_0$ (\cref{fig:ceiling}). Cells
the robot never passes over stay at the recorded floor height~$h_{\mathrm{floor}}$, except
for a one-cell margin around planted feet, and a running median clamped beneath the result
removes narrow notches left by grazing toes without rounding step edges. On a stair ascent
the rays stop at the stance soles, so $h_0$ already holds every tread at stance height.

Setting $h_0$ this way places elevated treads and seats at their kinematic height without
initial interpenetration at iteration~$0$, raising the initial foot-support rate from $0.47$
to $0.80$ on our multi-contact development clip. Retargeting offsets and compliant PD
contact let the simulated robot settle slightly below raw poses, so closed-loop edits
(\cref{eq:edit}) may raise $h_0$ up to $\eta{=}5$\,cm.

\subsection{Contact Mimic Rewards and Shape Regularization}
\label{sec:objective}\label{sec:sac}
The geometric ceiling $h_0$ meets the posed reference body at each cell's lowest swept
height regardless of whether the body receives ground reaction forces there or passes
through free space. Realizing \emph{terrain-as-mimic} therefore drives sculpting with
contact-mimic rewards in the cycle execution return $R(h_k,\pi_\theta)$ learned by the
SAC critic, supplemented on the actor with auxiliary shape priors that lower
uncontacted ceiling cells and suppress artifacts.

\paragraph{Cycle execution reward $R(h_k,\pi_\theta)$.}
Contacts are non-differentiable in~$h_k$, so each surface is scored at cycle
close from $\pi_\theta$ rollouts for $Q(s^{\mathrm{ter}}_k,u_k)$ and selection
(\cref{sec:population}),
\begin{equation}\label{eq:score}
  R(h_k,\pi_\theta)=\frac{\bar L}{L_{\max}}-\lambda_f\,\bar F-\frac{\bar e_z}{\sigma}
    +\lambda_c\,\bar c.
\end{equation}
Averaged over the episodes completed during cycle~$k$, the first three terms evaluate
closed-loop tracking. Specifically, $\bar L/L_{\max}\in[0,1]$ rewards mean episode duration
normalized by the $L_{\max}{=}500$-step horizon, whereas $\bar F\in[0,1]$ penalizes the
fraction of episodes cut short by a tracking failure so late falls carry the same penalty as
early falls, and $\bar e_z/\sigma$ penalizes mean absolute vertical torso-anchor error
normalized by the $\sigma{=}0.25$\,m termination threshold.

\paragraph{Contact mimic rewards.}
Maximizing tracking terms alone admits degenerate solutions that raise the surface
uniformly beneath an unstable policy to prevent falls and prop up torso height.
The fourth term, \emph{contact mimic} $\bar c\in[0,1]$, ties
terrain height to the motion by requiring simulated contacts to occur at the elevations
implied by the reference poses. For each solver-reported contact point on link~$b$, we
measure the vertical height error $\Delta z$ between its simulated position and the same
link-local point posed on the reference frame and score it with an exponential kernel
$\exp(-|\Delta z|/h_c)$ ($h_c{=}0.1$\,m), smoothly down-weighting contacts with large
horizontal offset so controller drift is not blamed on the terrain.
We average $\bar c$ only over steps with terrain contact, omitting airborne steps, and
aggregate per-contact scores onto the grid to form $A_k$ (\cref{sec:editor}).%
\begin{table}[t]
  \caption{Surface-roughness losses $\mathcal{L}_{\mathrm{turn}}$, $\mathcal{L}_{\mathrm{pit}}$,
    and $\mathcal{L}_{\mathrm{rough}}=\mathcal{L}_{\mathrm{turn}}+\mathcal{L}_{\mathrm{pit}}$
    on 1D terrain profiles extruded across the $126\times126$ ($0.08$\,m) grid
    ($\sigma{=}0.25$\,m, closing half-width $0.30$\,m).}
  \label{tab:rough}
  \centering
  \footnotesize
  \setlength{\tabcolsep}{4pt}
  \begin{tabularx}{\linewidth}{@{}>{\raggedright\arraybackslash}Xrrr@{}}
    \toprule
    Terrain profile & $\mathcal{L}_{\mathrm{turn}}$ & $\mathcal{L}_{\mathrm{pit}}$ & $\mathcal{L}_{\mathrm{rough}}$ \\
    \midrule
    \multicolumn{4}{@{}l}{\emph{Desired terrain}} \\
    Planar slope ($30\%$ grade)                  & $0$     & $0$     & $\mathbf{0}$ \\
    Sharp step ($1.5$\,m) / staircase            & $0$     & $0$     & $\mathbf{0}$ \\
    Wide cleared wall ($2.0$\,m wide, $1.0$\,m)  & $0$     & $0$     & $\mathbf{0}$ \\
    \midrule
    \multicolumn{4}{@{}l}{\emph{Undesired artifacts}} \\
    Raised wave ($5$\,cm cell-wise)              & $0.200$ & $0$     & $\mathbf{0.200}$ \\
    Narrow flat trench ($0.4$\,m wide, $0.4$\,m) & $0$     & $0.063$ & $\mathbf{0.063}$ \\
    Carved wave ($5$\,cm cell-wise)              & $0.200$ & $0.100$ & $\mathbf{0.300}$ \\
    \bottomrule
  \end{tabularx}
\end{table}

\paragraph{Auxiliary shape regularization $\mathcal{L}_{\mathrm{shape}}(h_k)$.}
\label{sec:bounds}
Execution reward alone leaves unvisited or contact-neutral cells underconstrained, yet
folding small per-cycle geometric penalties into the scalar critic reward swamps them in
rollout variance. To supplement SAC without polluting critic targets, we evaluate an
explicit differentiable shape loss $\mathcal{L}_{\mathrm{shape}}(h_{k+1})$ on the retained
surface $h_{k+1}=\mathcal{E}(h_k,\bar u_k)$ (\cref{eq:edit}) and add it directly to the
standard SAC actor objective alongside auxiliary reset-contact and step-size regularizers.
Scaled by~$\sigma$, $\mathcal{L}_{\mathrm{shape}}(h_k)$ sums three non-negative terms,
\begin{equation}\label{eq:shape}
  \begin{aligned}
    \mathcal{L}_{\mathrm{shape}}(h_k)={}&\lambda_{\mathrm{elev}}\,\mathcal{L}_{\mathrm{elev}}(h_k)
      +\lambda_{\mathrm{bnd}}\,\mathcal{L}_{\mathrm{bnd}}(h_k)\\
      &+\lambda_{\mathrm{rough}}\,\mathcal{L}_{\mathrm{rough}}(h_k).
  \end{aligned}
\end{equation}
Here $\mathcal{L}_{\mathrm{elev}}(h_k)$ is the normalized grid-mean height whose gradient
pulls non-supporting ceiling elevations toward the floor, and
$\mathcal{L}_{\mathrm{bnd}}(h_k)$ quadratically penalizes heights outside
$[\underline{h}_0,h_0]$ to tolerate centimetre-level support calibration while suppressing
large protrusions or trenches.

Unlike second-difference curvature, which penalizes sharp steps as heavily as waves,
$\mathcal{L}_{\mathrm{rough}}(h_k)=\mathcal{L}_{\mathrm{turn}}(h_k)
+\mathcal{L}_{\mathrm{pit}}(h_k)$ combines axial peak-or-dip turning across all interior
consecutive $3$-cell row and column triplets $(z^-,z,z^+)\in\mathcal{T}$ with pit depth
beneath a stop-gradient morphological closing $\operatorname{cl}(h_k)$ of half-width
$0.30$\,m ($9\times9$),
\begingroup\allowdisplaybreaks[1]
\begin{subequations}\label{eq:rough}
\begin{align}
  \mathcal{L}_{\mathrm{turn}}(h_k) &= \frac{1}{N(N{-}2)\sigma}\sum_{\mathcal{T}}
    \Bigl(\bigl[z-\max(z^-,z^+)\bigr]_+ \nonumber\\
    &\hspace{2.2em} + \bigl[\min(z^-,z^+)-z\bigr]_+\Bigr), \label{eq:turn}\\
  \mathcal{L}_{\mathrm{pit}}(h_k) &= \frac{1}{N^2\sigma}\!\sum_{i,j}\!
    \bigl[\min\bigl(\operatorname{cl}_{ij}(h_k),h_{0,ij}\bigr){-}h_{k,ij}\bigr]_+, \label{eq:pit}
\end{align}
\end{subequations}
\endgroup
where $[z]_+=\max(z,0)$. Both losses vanish on planar slopes, monotone staircases, and wide
cleared walls (\cref{tab:rough}), while $\mathcal{L}_{\mathrm{turn}}$ penalizes raised waves
above $h_0$ missed by $\mathcal{L}_{\mathrm{pit}}$ ($0.200$ vs.\ $0$) and
$\mathcal{L}_{\mathrm{pit}}$ penalizes multi-cell flat-bottomed trenches whose monotone
walls evade $\mathcal{L}_{\mathrm{turn}}$ ($0.063$ vs.\ $0$, five times the $0.013$
reduction in $\lambda_{\mathrm{elev}}\mathcal{L}_{\mathrm{elev}}$).

\subsection{Co-Optimization Workflow}
\label{sec:population}\label{sec:sim}\label{sec:planning}

\begin{figure*}[t]
  \centering
  \includegraphics[width=0.80\linewidth]{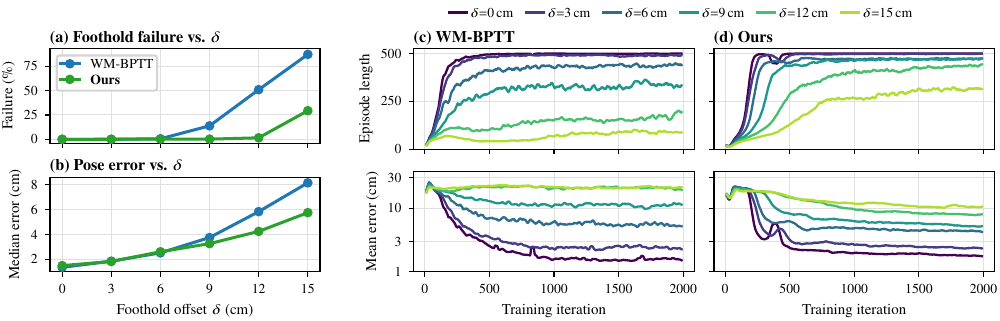}
  \caption{Controlled stair-foothold perturbation across $\delta\in\{0,3,6,9,12,15\}$\,cm
    ($2{,}000$ iterations on frozen terrain). Panels~(a,~b) show uniform-start mean foothold
    failure and median pose error, and (c,~d) show training curves for WM-BPTT and
    \textbf{Ours}.}
  \label{fig:stairs-perturb}
\end{figure*}

Optimizing a single terrain in place yields only one sculpting transition every $I{=}96$
imitation iterations, while $\pi_\theta$'s ongoing improvement raises raw $R$ over time
even on an unchanged surface. We instead partition the $8{,}192$ parallel environments
across $K{=}64$ heightfield tiles of $128$ environments each, pairing $8$ unedited
reference tiles holding $h^{\mathrm{inc}}$ (initialized to $h_0$) with $56$ exploration
tiles ($7$ per reference) that test cut-biased edits $\mathcal{E}(h_k,u_k)$. Giving each
tile $128$ environments covers the reference clip within one cycle to populate
$s^{\mathrm{ter}}_k$, while $56$ parallel exploration tiles multiply sculpting throughput
and train the shared perceptive policy $\pi_\theta$ across diverse nearby geometries.

Co-optimization over this population alternates step-level adaptation under $V(s_t)$ with
cycle-boundary sculpting under $Q(s^{\mathrm{ter}}_k,u_k)$. Within each $96$-iteration
cycle, installed surfaces remain fixed while $\pi_\theta$ and $V(s_t)$ update on-policy
across all $64$ tiles and accumulate contact statistics into $s^{\mathrm{ter}}_{k+1}$.
Before sculpting converges, nominal footholds can lack support. Joint commands that
lead to a stable alternative can receive positive advantage in PPO rollouts and
become more likely under the policy (\cref{sec:exp-planner}).

At each cycle boundary, subtracting the paired reference tile's execution return
$R(h^{\mathrm{inc}},\pi_\theta)$ from $R(\mathcal{E}(h_k,u_k),\pi_\theta)$ and zeroing
differences within rollout noise isolates the terrain edit's effect as a scalar critic
reward $r_k$ free of $\mathcal{L}_{\mathrm{shape}}$. The $56$ exploration tiles then retain
their mean-action surfaces $h_{k+1}=\mathcal{E}(h_k,\bar u_k)$ and append transitions
$(s^{\mathrm{ter}}_k,u_k,r_k,s^{\mathrm{ter}}_{k+1})$ to the replay buffer, where
off-policy SAC fits $Q(s^{\mathrm{ter}}_k,u_k)$ to $r_k$ and updates $\pi_\phi$ with the
analytic gradient of $\mathcal{L}_{\mathrm{shape}}(h_{k+1})$ (\cref{sec:bounds}). To
prevent failed cuts or diverging tiles from destabilizing the shared policy $\pi_\theta$,
any tile trailing its own best $R-\mathcal{L}_{\mathrm{shape}}$ for two consecutive cycles
reverts to that best surface, and every $10$ cycles ($960$ iterations) all $64$ tiles
consolidate onto the highest-objective surface so far as the new $h^{\mathrm{inc}}$.
\Cref{app:training} details the complete training procedure, with hyperparameters and
simulation engineering in \cref{app:hyper,app:engineering}.

\section{Dual-Network Co-Optimization Analysis}
\label{sec:exp}\label{sec:method-protocol}

We evaluate \method{} across PPO vs.\ world-model tracking
(\cref{sec:exp-planner}), closed-loop vs.\ kinematic terrains (\cref{sec:exp-kinematic}),
sculptor designs (\cref{sec:exp-sculptor}), and contact labels on held-out 3D scans
(\cref{sec:exp-egohtr}).

\begin{figure}[tb]
  \centering
  \includegraphics{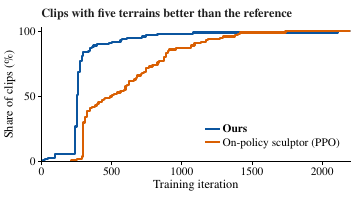}
  \caption{Cold-start terrain discovery of \textbf{Ours} and an on-policy sculptor (PPO) on
    $100$ paired LAFAN1 clips, showing the fraction of clips whose sculptor has found
    at least five terrain edits improving tracking beyond rollout noise
    (\cref{sec:population}).}
  \label{fig:sculptor-speed}
\end{figure}

\begin{table}[tb]
  \caption{Terrain--policy co-optimization on $53$ LAFAN1 obstacle clips
    ($12{,}000$-iteration budget). A run converges if its final mean episode length
    reaches $450/500$ steps, and completion is the fraction of $256$ first-second rollouts
    that finish.}
  \label{tab:wm-coopt}
  \centering
  \footnotesize
  \setlength{\tabcolsep}{0pt}
  \begin{tabular*}{\linewidth}{@{\extracolsep{\fill}}lrr@{}}
    \toprule
    Co-optimization metric & WM-BPTT & \textbf{Ours} \\
    \midrule
    Converged runs (\%)              & $58.5$ & $\mathbf{98.1}$ \\
    Full-clip completion (\%, med.)  & $84.1$ & $\mathbf{99.2}$ \\
    Tracking failures / $10^3$ steps & $0.59$ & $\mathbf{0.08}$ \\
    \bottomrule
  \end{tabular*}
\end{table}

\begin{figure}[tb]
  \centering
  \includegraphics[width=\linewidth]{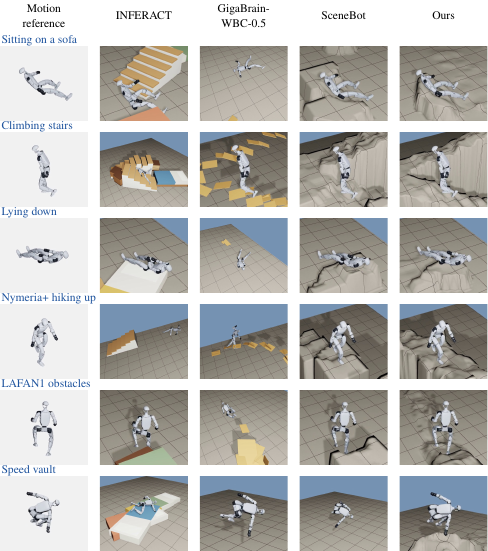}
  \caption{Rollouts at one motion frame per row, pairing our tracker with INFERACT
    \citep{li2024physics} (our box library, drawn as furniture), GigaBrain-WBC-0.5
    \citep{cheng2026gigabrain}, SceneBot \citep{chen2026scenebot}, and ours. Fallen robots
    are zoomed out.}
  \label{fig:gallery}
\end{figure}

\subsection{PPO and World-Model Tracking}
\label{sec:exp-planner}

Short-horizon analytic policy gradients can pass through differentiable simulation
\citep{xu2022shac} or a learned dynamics model as in SuperMimic
\citep{liao2026supermimic}, which scales flat-ground tracking including jumps and flips.
\Cref{tab:wm-coopt} compares two systems on LAFAN1 obstacles
\citep{harvey2020lafan1}. Our world-model variant (\emph{WM-BPTT}) uses model
gradients for tracking and terrain edits. \method{} uses PPO and SAC, respectively.
Both use the same scans, rewards, terminations, and paired terrain selection, with
GAE targets for their critics. WM-BPTT tracks compatible surfaces slightly more
accurately when it converges, but often stalls or collapses when nominal footholds
lack support and adaptive resets favor blocked steps. \method{} converges on more
clips and completes them more reliably.

Both trackers explore in physics simulation. PPO weights sampled joint commands by
advantages estimated from simulated rewards and critic values, without online
sampling-based MPC \citep{xue2025dialmpc}. WM-BPTT fits dynamics to the collected
transitions and draws rollout starts from replay. It updates the actor by
differentiating predicted return through $32$ mean-action model steps plus
terminal value.

To compare trackers on identical terrain, we freeze the stairs, shift footholds
vertically by $\pm\delta$, and retrain from scratch. At small offsets both retain
nominal footholds with near-zero failure and WM-BPTT has slightly lower pose error
(\cref{fig:stairs-perturb}a,b). At larger offsets WM-BPTT often falls on displaced
treads. PPO uses earlier treads or adjacent ground, changing at least one nominal
foot placement in all $966$ successful crossings at $\delta{=}12$\,cm
(\cref{fig:stairs-perturb}a,c,d, \cref{app:exp-planner}).

\subsection{Heightfield Sculpting vs.\ Scene Baselines}
\label{sec:exp-kinematic}\label{sec:exp-exec}\label{sec:method-baselines}

Prior motion-only pipelines either fit static surfaces to low-velocity contacts
\citep{chen2026scenebot,cheng2026gigabrain}, with the swept envelope~$h_0$
(\cref{sec:ceiling}) as a kinematic bound, or place rigid library objects on the floor in
closed loop \citep{li2024physics}. On scan-free clips, survival alone cannot validate
geometry, as anchor-relative rewards (\cref{sec:tracker}) let a policy finish despite
uncarved passes or contact sinkage. On LAFAN1 obstacles (\cref{tab:ceiling}), static $h_0$
matches far fewer contacts and rides lower than \method{}, while GigaBrain-WBC-0.5 and
SceneBot degrade further as velocity rules miss dynamic contacts and reaction forces.
Discrete box layout (INFERACT) scores lowest, as fixed templates miss obstacle profiles
and periodic layout resets disrupt policy adaptation. Sculpting a continuous surface from
closed-loop physical feedback is thus needed to separate supports from close passes and
calibrate height.

\begin{table}[tb]
  \caption{LAFAN1 obstacle grounding ($21$\,min, clip medians).}
  \label{tab:ceiling}
  \centering
  \footnotesize
  \setlength{\tabcolsep}{0pt}
  \begin{tabular*}{\linewidth}{@{\extracolsep{\fill}}lrrr@{}}
    \toprule
    Method & \hstack[r]{Contact mimic\\(\%)} & \hstack[r]{Torso error\\(cm)}
      & \hstack[r]{Torso sink\\(cm)} \\
    \midrule
    \multicolumn{4}{@{}l}{\emph{Prior scene baselines and kinematic bound}} \\
    INFERACT         & $15.4$ & $14.3$ & $11.0$ \\
    GigaBrain-WBC-0.5 & $42.3$ & $8.2$ & $6.8$ \\
    SceneBot         & $52.9$ & $6.1$ & $5.9$ \\
    Ceiling $h_0$    & $63.0$ & $4.4$ & $4.2$ \\
    \midrule
    \multicolumn{4}{@{}l}{\emph{Closed-loop sculpting (from $h_0$)}} \\
    CEM              & $57.3$ & $5.1$ & $5.1$ \\
    Shape in critic  & $68.0$ & $3.7$ & $3.0$ \\
    \textbf{Ours}    & $\mathbf{73.7}$ & $\mathbf{2.9}$ & $\mathbf{2.1}$ \\
    \bottomrule
  \end{tabular*}
\end{table}

\begin{figure}[tb]
  \centering
  \includegraphics[width=0.92\linewidth]{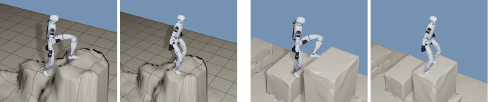}
  \caption{EgoHTR box-step climb ($1.9$\,s apart), showing \method{} on inferred terrain
    (left) and the reference in the held-out scan (right).
    Here and in \Cref{fig:egohtr-climb-full}, we render
    \href{https://huggingface.co/datasets/leggedrobotics/egohtr}{EgoHTR} scene scans
    \citep{brandes2026egohtr}, licensed under \href{https://creativecommons.org/licenses/by/4.0/}{CC BY 4.0}.}
  \label{fig:egohtr-climb}
\end{figure}

\subsection{Sculptor Optimization and Regularization}
\label{sec:exp-sculptor}\label{sec:method-ablation}

Sculpting a dense heightfield within $\sim\!125$ non-differentiable cycles requires
spatial state conditioning, decoupled regularization, and sample reuse. Directly
optimizing the $3{\times}84{\times}84$ brush space without a policy via parallel
cross-entropy search (CEM) relies on a single scalar return without spatial contact maps
$s^{\mathrm{ter}}_k$ or analytic shape gradients, stalling near $h_0$ while unguided
perturbations drop contact mimic below static $h_0$ (\cref{tab:ceiling}). A
state-conditioned SAC sculptor also underperforms when shape priors enter scalar $r_k$
(\emph{Shape in critic}), as rollout variance buries per-cell gradients and
under-calibrates support, whereas routing $\mathcal{L}_{\mathrm{shape}}$ analytically
through~$\mathcal{E}$ to the actor (\cref{sec:bounds}) keeps per-cell supervision while
the critic models execution alone. Against a matched on-policy PPO sculptor
(\cref{fig:sculptor-speed,app:experiments}), off-policy replay roughly halves cold-start
terrain discovery time. Spatial state conditioning, critic--actor decoupling, and
off-policy replay thus jointly enable sample-efficient sculpting.

\subsection{Contact Validation on Scanned Scenes}
\label{sec:exp-egohtr}

\begin{table}[tb]
  \caption{Rollout-voted contact accuracy against held-out 3D scans on $22.5$\,min of
    EgoHTR motion, reporting missed supports, F1, and median terrain-to-scan height error.}
  \label{tab:egohtr-contact}
  \centering
  \footnotesize
  \setlength{\tabcolsep}{0pt}
  \begin{tabular*}{\linewidth}{@{\extracolsep{\fill}}lcccc@{}}
    \toprule
    Scene & \hstack{Duration\\(min)} & \hstack{Missed\\(\%)} & F1 & \hstack{Height error\\(cm)} \\
    \midrule
    Walking, standing      & $\phantom{0}5.2$ & $0.5$ & $0.992$ & $3.1$ \\
    Stairs, parkour        & $\phantom{0}7.8$ & $0.7$ & $0.976$ & $1.8$ \\
    High objects           & $\phantom{0}4.2$ & $1.9$ & $0.960$ & $1.8$ \\
    Cliffs, rocks, stones  & $\phantom{0}5.3$ & $1.1$ & $0.965$ & $2.3$ \\
    \midrule
    All                    & $22.5$ & $0.9$ & $0.975$ & $2.3$ \\
    \bottomrule
  \end{tabular*}
\end{table}

To test whether motion-only co-optimization recovers physical contacts in real 3D scenes,
we run \method{} on EgoHTR recordings \citep{brandes2026egohtr} while holding out their
paired laser scans (\cref{fig:egohtr-climb}). Across walkways, stairs, high platforms, and
rocky paths (\cref{tab:egohtr-contact,app:egohtr}), rollout-voted contacts
(\cref{sec:data}) capture nearly all scanned foot supports with high F1 and close
elevation agreement under the simulated feet. Co-optimizing terrain and control
from kinematics alone thus recovers accurate support timing and height in real scenes,
validating the supervision used for \data{} and \adapt{}.

\section{Scaling to \data{} and \adapt{}}
\label{sec:data}\label{sec:labels}\label{sec:adapt}

\method{} infers terrain without scans or templates from $74.2$\,h of Nymeria+
motion tracks alone \citep{ma2024nymeria,detone2026nymeriaplus} and $1.0$\,h of LAFAN1
obstacles \citep{harvey2020lafan1} ($75.2$\,h in \data{}), compared in
\cref{tab:motion-scale} with scanned, retargeted, and GRAIL data
\citep{xie2026grail,nvidia2026graildata}.

\begin{table}[tb]
  \caption{Motion duration by source.}
  \label{tab:motion-scale}
  \centering
  \footnotesize
  \setlength{\tabcolsep}{2pt}
  \begin{tabularx}{\linewidth}{@{}l>{\raggedright\arraybackslash}Xr@{}}
    \toprule
    Source & Motion measured & Hours \\
    \midrule
    EgoHTR \citep{brandes2026egohtr} & Scanned motion & $1.37$ \\
    OmniRetarget \citep{yang2025omniretarget} & Retargeted clips & $8.4$ \\
    GRAIL \citep{nvidia2026graildata} & Synthetic G1 ($25$\,Hz) & $93.3$ \\
      & \hspace{0.4em}Curbs, slopes, stairs & $73.6$ \\
    \midrule
    \textbf{\data{} (ours)} & Motion-only clips & $75.2$ \\
    \bottomrule
  \end{tabularx}
\end{table}

\Cref{fig:gallery} compares our specialists on inferred terrain with the same tracker
trained under three prior motion-to-scene methods. Since wearable drift
misleads height thresholds, we replay each specialist from random starts and vote over
aligned valid steps, excluding post-reset settling, pre-failure windows, and pose or
height outliers to extract $10$-part contact schedules $\cstar$ ($<0.9\%$ cross-seed
foot disagreement, \cref{app:data}).

\begin{figure}[tb]
  \centering
  \includegraphics[width=0.85\linewidth]{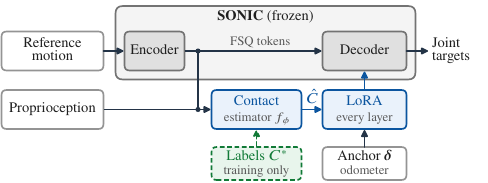}
  \caption{\adapt{}. Frozen SONIC (grey) pairs with rank-$16$ decoder LoRAs (blue). Only
    the first LoRA reads $\mathbf{u}_t$, combining predicted contacts
    $\chat_{t:t+9}$ (via $f_\phi$ from $\cstar$) and anchor $\boldsymbol\delta_t$.}
  \label{fig:sonic-t}
\end{figure}

\begin{table}[t]
  \caption{Tracking success (\%) on $1.9$\,h of held-out motion in $10$-s windows
    ($\pm$95\% CI half-width) after $10{,}000$ training iterations.}
  \label{tab:utility}
  \centering
  \footnotesize
  \setlength{\tabcolsep}{0pt}
  \begin{tabular*}{\linewidth}{@{\extracolsep{\fill}}lccc@{}}
    \toprule
    Policy & Overall & Terrain & Flat+mix \\
    \midrule
    Frozen SONIC          & 51.7 {\scriptsize$\pm$6.3} & 8.1 {\scriptsize$\pm$4.9} & 79.7 {\scriptsize$\pm$5.4} \\
    \adapt{} (uncond.)    & 74.3 {\scriptsize$\pm$5.1} & 49.7 {\scriptsize$\pm$9.1} & 89.6 {\scriptsize$\pm$4.0} \\
    \textbf{Ours} (\adapt{}) & \textbf{89.3} {\scriptsize$\pm$3.5} & \textbf{92.2} {\scriptsize$\pm$5.6} & \textbf{89.8} {\scriptsize$\pm$4.0} \\
    \bottomrule
  \end{tabular*}
\end{table}

We adapt frozen SONIC \citep{luo2026sonic} via rank-$16$ decoder LoRA \citep{hu2022lora}
conditioned on predicted contacts and pelvis anchor (\cref{fig:sonic-t}). On $14.7$\,h of
\data{}, unconditioned LoRA lifts held-out terrain success from $8.1\%$ to $49.7\%$, and
$\mathbf{u}_t$ conditioning reaches $92.2\%$ at matched budget (\cref{tab:utility}) and
$96.7\%$ at $16{,}500$ iterations ($90.3\%$ with learned odometry, \cref{app:downstream}).

\section{Conclusion}
\label{sec:limits}\label{sec:conclusion}

Under two-timescale \emph{terrain-as-mimic} co-optimization, PPO learns to place the foot
on earlier treads or adjacent ground under support mismatch where our WM-BPTT tracker
often loses balance, while decoupling execution-only critic targets from analytic actor
shape priors under off-policy SAC replay improves sample efficiency. Built on this design,
\method{} grounds $75.2$\,h of motion-only archives in physics without scans or
templates, building \data{} and adapting SONIC into \adapt{}.

Two limitations guide future work. First, a static, single-valued $2.5$D heightfield
precludes under-table clearance, stacked floors, and movable supports, calling for dynamic
multi-layer 3D geometry. Second, \method{} focuses on the physical executability of
embodied motion under spatially discrete contacts, so its sculpted terrains capture local
support and clearance rather than the full distribution of real-world scenes.
Terrain-interactive motor skill, however, can be decoupled from scene perception via a
contact interface \citep{he2026acp}. Future work will pair our contact-guided controller
with a standalone perception-and-planning module driven by onboard visual observations.

\clearpage

\bibliography{mimic_t_gen}
\bibliographystyle{icml2026}

\appendix
\onecolumn

\section{Training Procedure}
\label{app:training}

\Cref{alg:training} summarizes how \method{} trains on one reference clip, and
\cref{tab:hyper-mtg} lists all hyperparameters.

\begin{algorithm}[htbp]
  \caption{\method{} co-optimization on one reference clip.}
  \label{alg:training}
  \begin{algorithmic}[1]
    \STATE Compute the geometric ceiling $h_0$ (\cref{sec:ceiling}), install it on all $64$ tiles, and set $h^{\mathrm{inc}}\gets h_0$.
    \STATE Initialize $\pi_\theta$, $V$, the sculptor actor $\pi_\phi$, its twin critics $Q$, and an empty replay buffer $\mathcal{D}$.
    \FOR{iteration $n=1,\dots,12{,}000$}
      \STATE Collect $24$ control steps in each of the $8{,}192$ environments and update $\pi_\theta$ and $V$ with PPO.
      \STATE Add every robot--terrain contact to the contact maps of its tile.
      \IF{the $96$-iteration cycle of a phase group closes (one group every $12$ iterations)}
        \STATE Score each exploration tile of the group against the group's reference tile and form $r_k$.
        \STATE Store $(s^{\mathrm{ter}}_k,u_k,r_k,s^{\mathrm{ter}}_{k+1})$ in $\mathcal{D}$ and take $32$ SAC steps on minibatches of $16$.
        \STATE Keep $h_{k+1}=\mathcal{E}(h_k,\bar u_k)$ on each exploration tile, or revert a tile that has trailed its best $R-\mathcal{L}_{\mathrm{shape}}$ for two cycles.
        \STATE Every $10$ cycles, restart the group's tiles from the best surface measured so far, which becomes $h^{\mathrm{inc}}$.
        \STATE Install $\mathcal{E}(h_{k+1},u_{k+1})$ with $u_{k+1}\sim\pi_\phi(\cdot\mid s^{\mathrm{ter}}_{k+1})$ on each exploration tile.
      \ENDIF
      \STATE Stop if the run has converged.
    \ENDFOR
    \STATE Consolidate all tiles and export the best surface as the terrain and $\pi_\theta$ as the specialist.
  \end{algorithmic}
\end{algorithm}

\paragraph{Imitation updates.}
In every iteration, each of the $8{,}192$ environments contributes $24$ control steps at
$50$\,Hz, and PPO updates $\pi_\theta$ and $V$ for $5$ epochs of $4$ minibatches
(\cref{tab:hyper-mtg}). Episodes start at reference frames drawn by the adaptive sampling of
\citet{liao2025beyondmimic}, which favors frames where episodes fail, and end at the $10$-s
limit or at a termination of \cref{sec:tracker}. The environments are split evenly over the
$64$ tiles, so one shared $\pi_\theta$ trains on all candidate surfaces at once.

\paragraph{Sculptor cycles.}
Tile $i$ belongs to phase group $i \bmod 8$. Tiles $0$--$7$ are the reference tiles of their
groups and hold $h^{\mathrm{inc}}$, and the other seven tiles of each group explore. The
$96$-iteration cycles of the eight groups close $12$ iterations apart, so the sculptor is
updated every $12$ iterations and each tile completes about $125$ cycles in $12{,}000$
iterations. When a group's cycle closes, the execution return of each exploration tile is
compared with that of the group's reference tile over the same cycle, and the difference
beyond the rollout-noise band becomes the critic reward $r_k$ (\cref{sec:population}). The
transitions enter a replay buffer of $256$, from which SAC takes $32$ gradient steps on
minibatches of $16$, with the twin critics regressing $r_k$ while the actor's loss adds the
analytic shape, reset-contact, and step-size terms evaluated on
$h_{k+1}=\mathcal{E}(h_k,\bar u_k)$. Each exploration tile then keeps $h_{k+1}$ and installs
a new sample $\mathcal{E}(h_{k+1},u_{k+1})$, whose target heights are biased toward cuts, for
its next cycle, while robots standing where the ground rose are lifted onto it
(\cref{app:engineering}).

\paragraph{Guard, consolidation, and stopping.}
A tile whose paired objective $R-\mathcal{L}_{\mathrm{shape}}$ stays below its own best for
two consecutive cycles reverts to that best surface and restarts its proposals from it.
Every $10$ cycles ($960$ iterations), each tile restarts at its next cycle boundary from the
best surface measured so far, which replaces $h^{\mathrm{inc}}$ only if an exploration tile
beat its reference. A run lasts at most $12{,}000$ iterations and stops early once
training has converged, specifically when at least $2{,}000$ iterations have elapsed, the mean
episode length over the last $500$ iterations reaches $98\%$ of its peak, and the mean reward,
contact mimic, body-position error, and anchor-position error each change between the last two
$500$-iteration windows by at most $2\%$, or by at most $0.01$ for contact mimic and $5$\,mm
for the two errors where these allowances are larger. A final consolidation puts the best
surface on every tile, exporting it as the terrain and the final $\pi_\theta$ as the
specialist.

\paragraph{Design checks.}
Single-clip development checks motivated this procedure. Assigning contacts to link origins
rather than solver contact points misplaced $69\%$ of contacts across cells. Two runs with
identical commands differed by $18.0$ vs.\ $13.8$ steps at the same early iteration, more
than a single edit changes, which is why each exploration tile is scored against a reference
tile from the same cycle. Swapping the policy altered $0.2$-s post-reset contact heights by
at most $0.9$\,cm. Omitting periodic consolidation increased the cross-tile spread of contact
mimic sixfold ($0.007\to0.045$ over $8.5$k iterations), whereas switching all tiles at once
set the tracker back for $\sim40$ iterations, which is why consolidation is staggered across
phase groups. Reusing archived cycles in an on-policy score-function sculptor drove its
policy loss from $0.03$ to $474$ within $200$ iterations.

\FloatBarrier
\section{Hyperparameters}
\label{app:hyper}

\Cref{tab:hyper-mtg} lists all simulation, terrain, tracker, objective, and sculptor settings
of \method{}.

\begin{table}[!htbp]
  \caption{Hyperparameters and architecture settings of \method{}.}
  \label{tab:hyper-mtg}
  \centering
  \footnotesize
  \setlength{\tabcolsep}{5pt}
  \renewcommand{\arraystretch}{1.13}
  \begin{tabularx}{\linewidth}{@{}l>{\raggedright\arraybackslash}p{4.8cm}>{\raggedright\arraybackslash}X@{}}
    \toprule
    Component & Parameter & Value \\
    \midrule
    Simulation & engine, solver & Newton, MuJoCo-Warp with MuJoCo contacts \\
               & physics step, control & 5\,ms, 50\,Hz (decimation 4) \\
               & environments, tiles, reference tiles & 8{,}192, 64 (128 environments each), 8, one per group \\
    \midrule
    Terrain    & tile, spacing, vertices, gap & $10\times10$\,m, 0.08\,m, $126\times126$, 2\,m \\
               & height window & lowest body $-1.0$\,m to highest body $+0.5$\,m \\
               & ceiling $h_0$ & swept URDF collision lower envelope, 1-cell margin ($<0.25$\,m/s), $0.56$\,m median clamped below the raw min \\
    \midrule
    Imitation $\pi_\theta$ & algorithm, networks & PPO, actor/critic MLP 512--256--128, ELU, input normalization \\
               & rollout, epochs, minibatches & 24 steps, 5, 4 \\
               & learning rate, schedule, KL target & $10^{-3}$, adaptive, 0.01 \\
               & $\gamma$, $\lambda$, clip, entropy, grad.\ clip, init.\ std & 0.99, 0.95, 0.2, 0.005, 1.0, 1.0 \\
               & episode, terminations & 10\,s (500 steps), anchor $z$ 0.25\,m, ori.\ 0.8, end-effector $z$ 0.25\,m, divergence \\
               & height scan & $17\times11$ points, 0.1\,m, $1.6\times1.0$\,m, no observation noise \\
               & iterations & at most 12{,}000, early stopping by the rule of \cref{app:training} \\
    \midrule
    Objective  & $\sigma$, $\lambda_f$, $\lambda_c$, $\lambda_{\mathrm{edit}}$ & 0.25\,m, 1, 1, 0.02 \\
               & shape weights $\lambda_{\mathrm{elev}}$, $\lambda_{\mathrm{bnd}}$, $\lambda_{\mathrm{rough}}$ & 0.2, 3, 1 ($\mathcal{L}_{\mathrm{rough}}=\mathcal{L}_{\mathrm{turn}}+\mathcal{L}_{\mathrm{pit}}$, \cref{eq:rough}) \\
               & $\mathcal{L}_{\mathrm{elev}}(h)$, $\mathcal{L}_{\mathrm{bnd}}(h)$ & grid mean of $h/\sigma$, grid mean of $([h-h_0]_+^2+[\underline{h}_0-h]_+^2)/\sigma^2$ ($\eta=5$\,cm) \\
               & $\mathcal{L}_{\mathrm{turn}}(h)$, $\mathcal{L}_{\mathrm{pit}}(h)$ & \cref{eq:turn} over interior axial triplets, \cref{eq:pit} with $9\times9$-cell closing ($0.30$\,m half-width, stop-grad on $\operatorname{cl}$) \\
               & contact mimic $\bar c$ & cutoffs $d_{\mathrm{near}}$ 0.2\,m, $d_{\mathrm{far}}$ 0.4\,m, height scale 0.1\,m, mean on contact steps \\
               & reset-contact $\mathcal{L}_{\mathrm{rst}}$ & first 10 steps (0.2\,s) after reset, $\ge3$ contacts/cell, dead band $\pm1$\,cm, $1-e^{-d/0.1\,\mathrm{m}}$, EMA 0.5, upward-miss weight $10\times$ \\
    \midrule
    Sculptor $\pi_\phi$ & action lattice, fields, resampling & 0.12\,m ($84\times84$), 3 ($\hat h,w,s$), bicubic $\tilde u=\operatorname{bicubic}(u_k)$ onto $126\times126$ grid \\
               & channel decoding & $\hat h=h_0+\eta\tanh\tilde u_1$ ($\tilde u_1\ge0$) or $h_0+(h_0-\underline{h}_0+\eta)\tanh\tilde u_1$ ($\tilde u_1<0$), $w=\rho\,\gamma(\tilde u_2)$, $s=\gamma(\tilde u_3)$ with smooth sigmoid $\gamma$ ($0.06$\,m edge) \\
               & blur std., reach $\rho$, step limit & 0.12\,m, swept region dilated by 0.24\,m, $\Delta_{\max}=5$\,cm ($\Delta_{\max}\tanh(\cdot/\Delta_{\max})$) \\
               & observation & 5 channels ($\tilde h_k$, $d_k$, $\tilde n_k$, $A_k$, and $e_k^+$, the positive part of the cell-mean signed height error of bodies touching the terrain one step after a reset, over $\sigma$), $A_k$ over cells with $\ge 8$ visits, clipped at $\pm3$ \\
               & networks & actor and twin critics with 2 conv layers ($3\times3$, 32, tanh) at terrain resolution, average-pooled to the lattice, actor $3\times3$ head (mean, log-std), critic head on pooled features concatenated with action, averaged to a scalar \\
    \midrule
    SAC        & buffer, batch, updates per cycle & 256, 16, 32 \\
               & $\gamma_\phi$, target rate $\tau$ & 0.9, 0.01 \\
               & initial $\alpha$, target entropy & 0.1, $-1$ per action dimension (log-probabilities averaged) \\
               & initial std, log-std range & 0.3, $[-5,1]$ \\
               & step size & $5\cdot10^{-3}\,\sigma/\mathrm{span}$, critics and $\alpha$ annealed $1/(1+k/30)$, fixed actor, grad.\ clip 1.0 \\
               & actor regularization & $(1-\gamma_\phi)^{-1}[\mathcal{L}_{\mathrm{shape}}(h_{k+1})+\lambda_c\mathcal{L}_{\mathrm{rst}}(h_{k+1})] + (\lambda_{\mathrm{edit}}/(N^2\sigma))\|h_{k+1}-h_k\|_1$ on $h_{k+1}=\mathcal{E}(h_k,\bar u_k)$ (horizon factor $10$), excluded from $r_k$ \\
               & exploration bias & target-height noise of the simulated $u_k$ samples above the mean at $1/4$ of lattice points ($3$-to-$1$ cut bias), keeping the actor distribution unchanged \\
    \midrule
    Population & cycle, tiles, phase groups & 96 iterations, 64 (8 reference, 56 exploration, 128 envs/tile), 8 (offset by 12) \\
               & critic reward $r_k$ & $\operatorname{sign}(\Delta R_k)\,[|\Delta R_k|-2\hat s]_+$ against group reference ($\Delta R_k = R(\mathcal{E}(h_k,u_k),\pi_\theta)-R(h^{\mathrm{inc}},\pi_\theta)$, $R$ only), $\hat s$ pooled from split halves over 8 cycles ($\ge0.003$) \\
               & guard & revert after 2 cycles below the tile's best $R-\mathcal{L}_{\mathrm{shape}}$ and restart proposals \\
               & consolidation & every 10 cycles (960 iterations) and at termination, all 64 tiles restart from the best measured surface, updating $h^{\mathrm{inc}}$ only if an exploration tile beat its reference \\
    \bottomrule
  \end{tabularx}
\end{table}

\FloatBarrier
\section{Simulation Engineering}
\label{app:engineering}

\paragraph{Heightfield contact limit.}
MuJoCo's heightfield collider keeps at most 50 contacts per primitive--heightfield pair and tests
two triangles for every cell a collision primitive covers, dropping contacts beyond the limit without
error. The G1's knee cylinders span 0.224\,m, which covers 72 triangles at 0.05-m spacing and
overflows as soon as the ground rises to meet the knee, as occurs on the initial ceiling
$h_0$. At 0.08\,m the largest primitive covers 32 triangles. We therefore compute the bound from the
robot's collision geometry and refuse smaller spacings. The coarser grid lowers the share of
frames on which the ceiling supports a foot within 1\,cm from 82.8\% to 76.8\% on our
multi-contact development clip.

\paragraph{Write path.}
Heightfields update in place on the GPU in Isaac Lab v3.0
with Newton v1.2.1 and MuJoCo-Warp v3.8.1
\citep{todorov2012mujoco,mujocowarp2025,newton2025}, and after every terrain update the
heightfield copies held by Newton and by MuJoCo-Warp are compared by hash. The reference motion's collision geometry penetrates the installed ceiling by zero
within the float32 encoding resolution, and one control step after reset every tracked body
is within 1--8\,mm of its reference height. Three tiles carrying identical terrain report
support rates of 0.7686, 0.7683 and 0.7685, so tiles in the shared scene do not leak contacts.
Wall-clock time scales linearly with tile count ($0.905+0.168\,n$\,s per iteration at
$2{,}048$ environments), and the full $64$-tile, $8{,}192$-environment configuration runs at
a median $2.5$\,s per iteration on an NVIDIA RTX~4090 including initialization
(\cref{sec:data}).

\paragraph{Warm-start accelerations.}
MuJoCo-Warp initializes each solve with the previous step's accelerations and writes the
result back, but environment resets do not clear this array. A non-finite acceleration
therefore survives the reset that should remove it and permanently corrupts that environment. In
one run, a single environment ended every episode in divergence from iteration 2{,}608 until
the end, and every reward curve became non-finite. Clearing the array for the reset environments
reduced non-finite rewards from 83{,}648 to 4 in a 12{,}000-iteration run, each recovered in
one step.

\paragraph{Lifting.}
No episode is reset at a cycle boundary, so a robot standing on ground that rises would
start the next step interpenetrating it. We lift each such robot by the largest rise under
its bodies, with observed lifts reaching 0.197\,m.

\paragraph{Height drift of inertial capture.}
On our multi-contact development clip, a foot that returns to the same horizontal position 6.4\,s later
reports the ground 0.28\,m higher, and the disagreement between revisits of the same stance
cell grows with the time between them (0.033\,m below 1\,s, 0.134\,m at 5--20\,s). This is
why the body-tracking terms of the tracker measure posture rather than absolute height
(\cref{sec:tracker}).

\FloatBarrier
\section{World-Model Co-Optimization and Stair Perturbation}
\label{app:exp-planner}

\paragraph{Variant settings and convergence.}
In \cref{sec:exp-planner,tab:wm-coopt}, the world-model co-optimization variant
(\emph{WM-BPTT}, with $H{=}32$-step BPTT, policy learning rate $2{\times}10^{-3}$, $8$ policy
updates and $96\!\to\!16$ model batches per iteration, $16$ tiles of $512$ environments, and
$40$-iteration cycles) is paired against \method{} across $53$ LAFAN1 obstacle clips
spanning $10$ sequences ($12{,}000$-iteration cap, bounded by the $13.3$\,h median wall-clock
of WM-BPTT). WM-BPTT reaches $\ge450/500$ final episode length on
$31/53$ clips ($58.5\%$, with $10$ never learning, $12$ collapsing after peaking, and $23/53$
stopping early at a median $12{,}000$ iterations or $13.3$\,h) versus $52/53$ ($98.1\%$, $0$
collapses, and $43/53$ stopping early at a median $4{,}481$ iterations or $3.1$\,h) for
\method{}. On the $31$ converged clips WM-BPTT reaches $2.4$ vs.\ $2.8$\,cm median pose
error ($1.4$ vs.\ $1.7$\,cm on the $25.8$-s stair climb), whereas across all $53$ clips
\method{} reaches $99.2\%$ vs.\ $84.1\%$ median frame-$0$ completion and $0.08$ vs.\ $0.59$
failures per $1{,}000$ steps ($100\%$ vs.\ $88.0\%$ median pass rate across $47$ deep
ledges, and $100\%$ vs.\ $0\%$ across the $8$ steepest $>20$\,cm ground-conflict ledges
where nominal reference footprints float above carved slopes and PPO laterally shifts
footsteps by $10$\,cm onto $+11$\,cm higher ground).

\paragraph{Auxiliary ablations.}
Controlled ablations of the variant confirm that
adding sole/collider clearances and contact flags lowers holdout one-step error by $<8\%$
($4\%$ on deep ledges), and mixing expert PPO rollouts into replay, per-trajectory
$10{\times}$-median gradient clipping, a binary cross-entropy termination head, or uniform
reset sampling ($0/233$ pass rate on $>20$\,cm conflicts, where $32$-step model--sim
response cosine drops from $0.54$ to $0.11$) all fail to produce reliable alternative foot placements.

\paragraph{Stair-foothold perturbation.}
To test foot placement under support mismatch, we freeze the converged stair terrain
and shift its $33$ interior footholds by $\pm\delta$ for
$\delta\in\{0,3,6,9,12,15\}$\,cm ($17$ raised and $16$ lowered). After $2{,}000$ iterations
of training from scratch (\cref{fig:stairs-perturb}), both policies achieve mean foothold failure
$\le0.5\%$ for $\delta\le6$\,cm ($1.4$ vs.\ $1.5$\,cm pose error at $\delta{=}0$\,cm). At
$\delta{=}9$\,cm, WM-BPTT jumps to $13.8\%$ vs.\ $0.2\%$ mean foothold failure (worse on all
$30$ shared footholds), $2\%$ ($6/247$) vs.\ $92\%$ frame-$0$ completion, and $2.40$ vs.\
$0.05$ failures per $1{,}000$ tracked steps, widening at $12$\,cm ($50.9\%$ vs.\ $1.5\%$)
and $15$\,cm ($87.1\%$ vs.\ $29.3\%$, with $8.1$ vs.\ $5.8$\,cm pose error). Loaded-foot
($>20$\,N) stance on raised nominal blocks drops in PPO from $100\%$ ($6$\,cm) to $78\%$,
$31\%$, and $13\%$ ($9,12,15$\,cm) onto prior treads ($22\%,58\%,46\%$) or adjacent unmoved
ground ($41\%$ at $15$\,cm). At $12$\,cm ($2\delta{=}24$\,cm $<25$\,cm limit), PPO abandons
nominal two-foot placement on $100\%$ of $966$ successful crossings ($1/967$ failures,
$0.1\%$, vs.\ $234/625$, $37\%$, for WM-BPTT, which plants on raised blocks $76\%$ of the
time, with start-bin concentration $0.09$--$0.27$ and $1$-s post-reset failure $13.8\%$ vs.\ $1.6\%$
at $9$\,cm).

\FloatBarrier
\section{Sculptor Baselines and Statistics}
\label{app:experiments}

\paragraph{Clips and measures.}
\Cref{tab:ceiling} uses $34$ LAFAN1 obstacle clips ($21$\,min), the first two
segments of each of the $17$ obstacle sequences. Every method trains one run per clip under
the same $12{,}000$-iteration budget and early-stopping rule (\cref{app:training}). Each
measure is averaged over the last $200$ training iterations of a run, and \cref{tab:ceiling}
reports the median over clips. Contact mimic is the reward $\bar c$ of \cref{sec:objective}
averaged over steps with terrain contact (counting rigid boxes placed by GigaBrain-WBC-0.5
and INFERACT alongside the floor in both the tracker's height scan and the contact readout),
torso error is the mean absolute height error of the tracked torso, and torso sink is its
mean signed height offset from the reference, negated. Paired differences over clips are
tested with the Wilcoxon signed-rank test, and discovery times with the sign test.

\paragraph{Baselines.}
All baselines keep the tracker, rewards, terminations, and budget of \method{}.
\emph{Ceiling $h_0$} trains $\pi_\theta$ alone on the fixed initial
surface $h_0$. \emph{GigaBrain-WBC-0.5} \citep{cheng2026gigabrain} trains $\pi_\theta$ alone
on rigid support boxes $1$\,cm thick above the flat floor, the scene its terrain annotation
produces. The annotator is not released, so we reimplement Section~3.3 of their report and
set the details it leaves open from their released annotations. Points spaced $2$\,cm apart
on the ankle-link meshes enter contact when their normal and tangential speeds fall below
$0.2$ and $0.3$\,m/s after a deceleration, contacts shorter than $10$ frames or within
$10$\,cm of the floor are discarded, a point stays only if a support $5$\,cm beyond it never
intersects the robot, and DBSCAN \citep{ester1996dbscan} groups the remaining points into
oriented boxes that grow sideways by at most $10$\,cm. On their $150$ benchmark clips, our
boxes reproduce theirs with a top-surface F1 of $0.78$. \emph{SceneBot}
\citep{chen2026scenebot} trains $\pi_\theta$ alone on the 2.5D elevation map of its
Algorithm~1, held fixed. A wrist, foot, or pelvis interacts while its speed stays below
$0.15$\,m/s and its acceleration below $2$\,m/s$^2$ for at least $0.1$\,s, the pelvis only
when more than $0.2$\,m from the feet. Each interaction raises a $1$\,m square platform to the
link's lowest collision point unless any link would sink more than $2$\,cm into it outside
that interaction, platforms within $0.1$\,m in height merge, and the map is carved down to
the swept lower envelope. Their report gives no values, so the speed threshold, platform
size, merge tolerance, and pelvis rule follow the released code of TIP
\citep{jiang2022tip}, on which their reconstruction builds, and the remaining values are
ours. \emph{INFERACT} \citep{li2024physics} trains $\pi_\theta$ jointly with its layout
generator, which places objects from our box library ($32$ objects) on the flat floor. For
every object slot the generator, trained with PPO, samples an object and its planar placement,
scored by the discounted tracking reward of episodes that touch an object, and every $504$
steps all environments restart among newly sampled layouts, exporting the generator's
deterministic mode-and-mean layout at the end of training. A contact value network regressed
on the tracker's critic marks frames where touching an object raises the value, and
complete-linkage clusters of their pelvis positions ($1$\,m, at least $0.2$\,s) set the number
of objects, at most eight, and the pose priors that anchor each placement. The paper gives no
values for the clustering, the prior weight, the placement range, or the generator's PPO
settings, so these are ours, and our tracker replaces their object-conditioned controller.
\emph{CEM} replaces the
learned actor with the cross-entropy method, where one diagonal Gaussian over the brush
action, shared by all exploration tiles and initialized at the untrained actor's output with
standard deviation $0.3$, draws the exploration samples, and once a sweep of cycles has
scored samples of the current distribution, the top $25\%$ refit its mean and spread with
smoothing $0.5$. Lacking a gradient, CEM scores a sample by its banded return gap to the
reference tile minus the shape and edit terms. \emph{Shape in critic} keeps the SAC sculptor but subtracts
the shape and edit terms from the critic reward $r_k$ instead of passing
$\mathcal{L}_{\mathrm{shape}}$ to the actor as an analytic gradient, which also removes the
reset-contact term. The \emph{PPO sculptor} trains the same actor on-policy
with PPO and keeps the rest of \method{}, including the analytic shape loss, but has no
replay.

\paragraph{Paired statistics on the $34$ clips.}
Against INFERACT, GigaBrain-WBC-0.5, SceneBot, Ceiling $h_0$, CEM, and Shape in critic,
\method{} reaches higher contact mimic on $34$, $32$, $32$, $32$, $34$, and $28$ of the $34$
clips ($p=1.2{\times}10^{-10}$, $2.2{\times}10^{-9}$, $5.0{\times}10^{-9}$, $1.6{\times}10^{-8}$,
$1.2{\times}10^{-10}$, and $1.0{\times}10^{-6}$), lower torso height error on $33$, $31$, $30$,
$33$, $33$, and $27$ clips ($p=2.3{\times}10^{-10}$, $1.3{\times}10^{-8}$, $2.0{\times}10^{-8}$,
$2.0{\times}10^{-7}$, $6.2{\times}10^{-8}$, and $5.4{\times}10^{-4}$), and keeps the torso less far
below the reference on $28$, $28$, $30$, $33$, $33$, and $27$ clips ($p=1.2{\times}10^{-6}$,
$1.6{\times}10^{-5}$, $2.5{\times}10^{-6}$, $1.5{\times}10^{-7}$, $2.4{\times}10^{-8}$, and
$3.7{\times}10^{-4}$). Although INFERACT continues sampling candidate layouts every $504$ steps
during training, deterministic full-clip replay on each method's exported scene (holding
INFERACT's final layout fixed without resampling or exploration noise) separates the same two
methods, with INFERACT and GigaBrain-WBC-0.5 completing medians of $0\%$ ($18/34$ clips at $0\%$,
reaching a median $6.8\%$ of the clip) and $20.5\%$ of their attempts against $99.0\%$ for
\method{} (lower on $27$ and $21$ clips, higher on $1$ and $7$, $p=5.1{\times}10^{-6}$ and
$7.3{\times}10^{-5}$), confirming that INFERACT's training-end gap reflects the exported layout
rather than online resampling. SceneBot, Ceiling $h_0$, CEM, and Shape in critic reach medians of
$99.2\%$, $99.8\%$, $99.6\%$, and $99.2\%$ ($p\ge0.64$).

\paragraph{Discovery speed and edits.}
CEM finds its fifth terrain beyond the noise band after a median $1{,}133$ iterations against
$251$ for \method{} (later on $33$ of $34$ clips, $p=4.1{\times}10^{-9}$) and its first after
$377$ against $239$ ($p=0.0033$), and on $21$ of the $34$ clips its exported terrain stays within
$0.2$\,mm of $h_0$. Shape in critic finds better terrain as fast as \method{} (median $251$
iterations for both, $p=0.49$) but edits less. Along the reference foot path it changes
$44\%$ against $64\%$ of cells by more than $1$\,cm ($33$ clips, $p=5.0{\times}10^{-6}$) and
raises them by $0.92$ against $1.46$\,cm on average ($p=8.3{\times}10^{-5}$), and it changes
$2.5\%$ against $5.1\%$ of the cells away from the path ($p=0.012$).

\paragraph{SAC versus PPO sculptor on $100$ clips.}
Comparing the off-policy SAC sculptor of \method{} against the on-policy PPO sculptor
under identical brush operators, $96$-iteration cycles, and seed~$0$ across all $100$
LAFAN1 obstacle segments shows comparable asymptotic survival at termination ($96.2\%$
vs.\ $95.9\%$, body error $4.34$ vs.\ $4.33$\,cm), whereas replay reuse ($72$ vs.\ $4$
sculptor updates by iteration $1{,}000$) yields physically superior exploration surfaces about
twice as early (\cref{fig:sculptor-speed}). Specifically, SAC reaches its fifth exploration
surface exceeding the noise band ($\Delta R_k{>}2\hat s$) at median iteration $251$ vs.\ $497$
($95/100$ clips, sign-test $p{=}1.3{\times}10^{-22}$, first $>\!\hat s$ win at $251$ vs.\
$359$, $p{=}2.6{\times}10^{-12}$), with no shape-only wins before iteration $1{,}000$ (median
$0\%$), higher first-consolidation paired scores ($0.080$ vs.\ $0.061$, Wilcoxon $p{=}0.033$),
deeper pass cuts ($-7.9$ vs.\ $-4.8$\,cm, $p{=}3.0{\times}10^{-9}$), and fewer far-field edits
($5.1\%$ vs.\ $6.4\%$, $p{=}2.9{\times}10^{-8}$).

\paragraph{Development clips.}
\Cref{tab:dev-clips} reports per-clip contact-structure measurements for three multi-contact
development clips trained with \method{} (resting supports $<0.5$\,m/s and aerial passes
$\ge2.0$\,m/s within $2$\,cm of $h_0$, plus all foot stances). On the $70.7$-s multi-contact
clip, $12{,}000$ iterations raise mean episode length from $22$ to $496/500$ steps ($98.6\%$
reach the time limit, $1.1\%$ end in end-effector and $0.4\%$ in anchor-position failures,
none diverge), lower body-position error from $0.210$ to $0.028$\,m, and reach $0.998$ support
rate with $0.91$ contact mimic, updating $h^{\mathrm{inc}}$ at iterations $959$, $2{,}879$, and
$10{,}559$. Squat reset-window hand contact drops from $39\%$ on $h_0$ to $4\%$ on the sculpted
terrain ($92\%\!\to\!63\%$ across full episodes), while vault hand-plant cells rise by
$+4.7$\,cm ($+3.7$\,cm across neighboring box cells).

\begin{table}[htbp]
  \caption{Per-clip contact-structure audit across three multi-contact development clips under
    each clip's trained specialist ($12{,}000$ iterations, where \emph{Reset} is the $0.2$-s
    post-reset window and \emph{Episode} is full uniform-start deterministic replay).}
  \label{tab:dev-clips}
  \centering
  \footnotesize
  \setlength{\tabcolsep}{4pt}
  \begin{tabularx}{\linewidth}{@{}>{\raggedright\arraybackslash}Xcccccc@{}}
    \toprule
    Clip (duration) & \hstack{$h_0$ sink\\(cm)} & \hstack{Stance lift\\(cm)}
                    & \hstack{Reset $\pm2$\,cm\\($h_0\!\to\!$Ours)}
                    & \hstack{Episode $\pm2$\,cm\\($h_0\!\to\!$Ours)}
                    & \hstack{Pass coll.\,R / E\\($h_0\!\to\!$Ours)}
                    & Replays completed \\
    \midrule
    Stair ascent ($25.8$\,s)          & $1.48$ & $+1.66$ & $73\%\!\to\!95\%$ & $81.5\%\!\to\!97.0\%$ & --- & $161/161$ \\
    Speed vault ($6.4$\,s)            & $1.79$ & $+2.13$ & $45\%\!\to\!83\%$ & $38.4\%\!\to\!83.2\%$ & $11\%\!\to\!4\%$ / $1.5\%\!\to\!0.3\%$ & $114/136$ \\
    Multi-contact session ($70.7$\,s) & $1.32$ & $+1.31$ & $67\%\!\to\!97\%$ & $61.0\%\!\to\!95.4\%$ & $0\%\!\to\!0\%$ / $1.2\%\!\to\!1.3\%$ & $425/426$ \\
    \bottomrule
  \end{tabularx}
\end{table}

\FloatBarrier
\section{Validation on Scanned Scenes}
\label{app:egohtr}

\paragraph{Data.}
EgoHTR \citep{brandes2026egohtr} records people moving through scenes that it also captures
with a laser scanner. We take $34$ clips from $31$ recordings ($22.5$\,min, $18$ optical and
$16$ inertial captures), chosen from the motion alone, and use the dataset's scene-agnostic
GMR retargeting to the G1 \citep{araujo2025gmr}, whose paired scans are scaled by the same
body-height ratio. \method{} runs on each clip from the motion alone, and each scan
is placed in its clip's frame with the offsets applied to the motion and is never observed
during training.

\paragraph{Supports.}
For every frame and foot, the lowest point of the foot's collision geometry is compared with
the highest scanned surface under its footprint (that point and $5$\,cm to each side). The
foot is supported when it lies less than $3$\,cm above that surface and its ankle moves
slower than $0.3$\,m/s, and free when it lies more than $6$\,cm above, while other frames are not
scored.

\paragraph{Contact labels.}
We replay each specialist about $250$ times from its clip's first second and vote foot
contacts as for \data{} (\cref{sec:labels}, $>1$\,N, majority of the valid executions). The
labels miss a support when they leave a supported frame free. The support height error is the
median absolute gap between the terrain and the scan under the simulated foot over supported
frames.

\begin{figure}[!htb]
  \centering
  \includegraphics{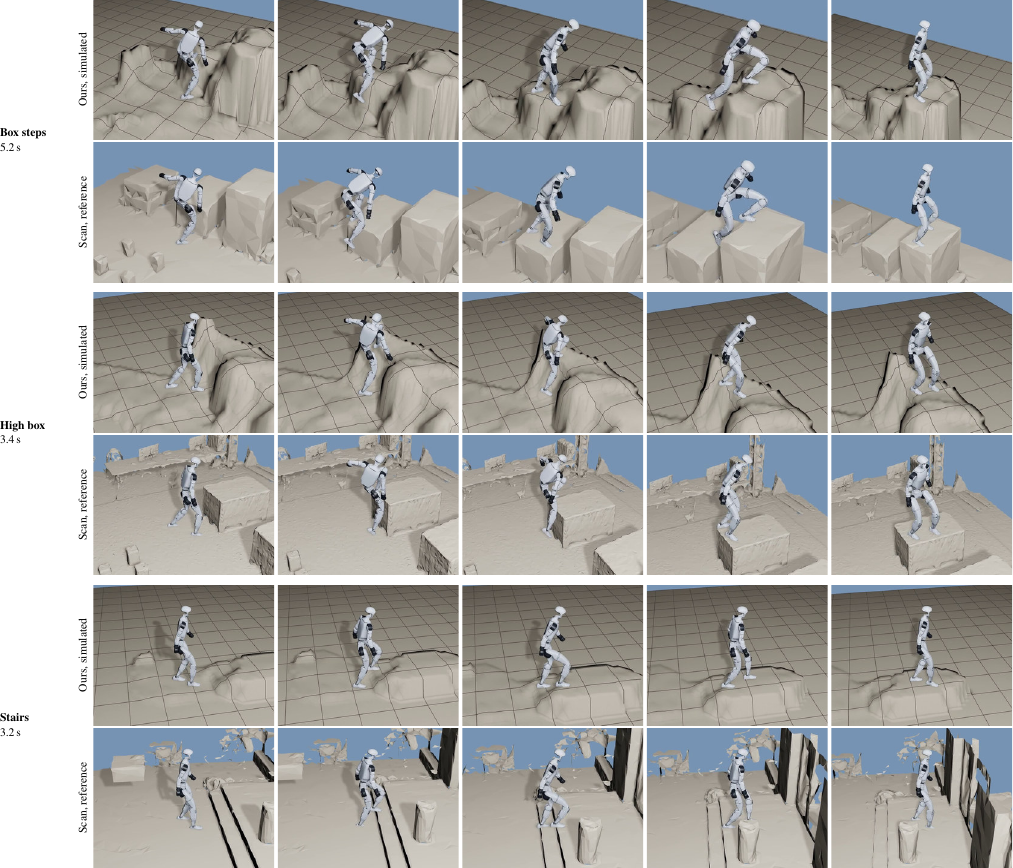}
  \caption{Three walk-ups in EgoHTR clips, each ending on top, five evenly spaced frames
    each (time span on the left). In each pair of rows, top shows the specialist simulated on the terrain \method{}
    inferred from motion alone, and bottom shows the reference motion drawn in the held-out
    laser scan of the scene (not simulated).}
  \label{fig:egohtr-climb-full}
\end{figure}

\paragraph{Results.}
\Cref{tab:egohtr-contact} groups the clips by EgoHTR activity into walking and standing
(including a narrow path), stairs and parkour, high objects, and cliffs, rocky paths, and
stepping stones. The reference itself stands
about $1.7$\,cm below the scan, which lowers all height errors alike.
\Cref{fig:egohtr-climb-full} shows three climbs.

\FloatBarrier
\section{\data{} Scale and Contact Extraction}
\label{app:data}

\paragraph{Corpus composition.}
\data{} comprises $75.2$\,h of sculpted motion across $6{,}289$ segments from $297$
recordings (\cref{sec:data}), combining $74.2$\,h of Nymeria+ recordings ($6{,}189$
segments across $280$ recordings, $18$ capture locations, and $11$ activity scripts) with
$1.0$\,h of LAFAN1 obstacle motion ($100$ segments across all $17$ obstacle sequences,
\cref{tab:nymeria-scale}). Sculpted terrain relief across Nymeria+ reaches up to $3.89$\,m,
with end-of-training medians across all $6{,}189$ segments of $0.994$ episode survival,
$1.4\%$ tracking-failure share, and $3.0$\,cm body-position error.

\paragraph{Recording segmentation and per-segment convergence.}
Long multi-room or two-story captures can wander across tens of metres or revisit the same
horizontal location at different floor levels, which would either stretch a fixed-size
heightfield to coarse spatial resolution or create vertical support contradictions on a
single-valued 2.5D surface. Before training, we greedily segment each retargeted recording
along time so every segment fits a single $10\times10$\,m tile at $0.08$\,m resolution and
remains free of multi-level height conflicts. First, a cut is placed whenever the horizontal
bounding box of the root trajectory exceeds $8$\,m on either axis, reserving a $2$\,m reach
margin inside the tile. Second, as frames are appended in half-second blocks, we track both
the segment's running swept lower envelope and the stationary underside body parts carrying
the center of mass at each non-airborne frame (ignoring free-fall flight phases such as
jumps), and place a cut whenever later frames pass beneath an earlier load-bearing support
and leave the body unsupported for $0.4$\,s, such as walking on a ground floor directly
beneath an upper-story room or staircase visited earlier. Stretches still longer than
$60$\,s are subdivided at their slowest-moving root frames, and every retained segment is
horizontally centered, ground-aligned at its lowest foot point, and rechecked on its final
grid (dropping fragments shorter than $5$\,s on Nymeria+ or $10$\,s on LAFAN1). On a single
GPU, $94.5\%$ of \method{} runs converge before the $12{,}000$-iteration cap, taking a
median of $3{,}221$ iterations or $2.3$\,h of wall-clock time per segment
($1.6$--$3.9$\,h from the 10th to 90th percentile).

\begin{table}[htbp]
  \caption{Motion duration and segment counts across Nymeria+ activity scripts and LAFAN1
    obstacle sequences in \data{}.}
  \label{tab:nymeria-scale}
  \centering
  \footnotesize
  \setlength{\tabcolsep}{8pt}
  \begin{tabular}{lrr}
    \toprule
    Source / Activity script & Hours & Segments \\
    \midrule
    Nymeria+ S7 Cooking              & 16.78 & 1{,}041 \\
    Nymeria+ S20 Party               & 13.17 &   824 \\
    Nymeria+ S10 Housekeeping        &  8.39 &   545 \\
    Nymeria+ S16 Simon says          &  7.96 &   610 \\
    Nymeria+ S12 Game night          &  7.79 &   495 \\
    Nymeria+ S18 Hike                &  6.10 & 1{,}527 \\
    Nymeria+ S2 Where is X           &  4.34 &   350 \\
    Nymeria+ S19 Fresh air           &  3.63 &   297 \\
    Nymeria+ S3 Welcome to my place  &  3.55 &   263 \\
    Nymeria+ S8 Having a meal        &  2.25 &   219 \\
    Nymeria+ S17 In the office       &  0.27 &    18 \\
    LAFAN1 Obstacles ($17$ sequences) &  1.00 &   100 \\
    \midrule
    Total                            & 75.25 & 6{,}289 \\
    \bottomrule
  \end{tabular}
\end{table}

\paragraph{Rollout-voted contact extraction.}
To label contacts without kinematic height thresholds that drift over long wearable
captures, each clip's trained specialist is replayed deterministically on its exported
terrain from uniformly random reference frames ($\sim\!32$ executions per frame over up to
$10$-s episodes). Contacts are tracked across $10$ body-part groups, namely left and right
feet (ankle pitch and roll links), left and right hands (wrist roll, pitch, and yaw links),
left and right knees, seat (pelvis and hip links), torso (waist and torso links), and left
and right arms (shoulder and elbow links). A simulated step enters the vote when it lies
past the $0.2$-s post-reset settling window, does not end the episode or fall within $1.0$\,s
before a tracking failure or clip end, stays within $0.10$\,m mean body-position error of
the posture-aligned reference, and keeps anchor height within $0.10$\,m of the frame-wise
median across executions so rare rollouts on a wrong stair tread are excluded without
relying on absolute reference height. A part registers contact when solver force on any of
its links exceeds $1$\,N. Taking the strict frame-wise majority across valid executions and
debouncing contact or flight runs shorter than $3$ frames ($60$\,ms) produces the binary
schedule $\cstar\in\{0,1\}^{T\times 10}$, paired with per-entry confidence weights
multiplying posterior majority agreement ($2\,\mathrm{P}(\mathrm{Beta}(k{+}1,n{-}k{+}1)>0.5)-1$
for $k$ contacts in $n$ valid executions) by the frame-wise control success rate. Across two
independent replay seeds, foot contact labels disagree on $0.07\%$--$0.90\%$ of frames, with
zero flips among the $10{,}618$ entries whose confidence weight is at least $0.9$.

\begin{figure}[htbp]
  \centering
  \includegraphics{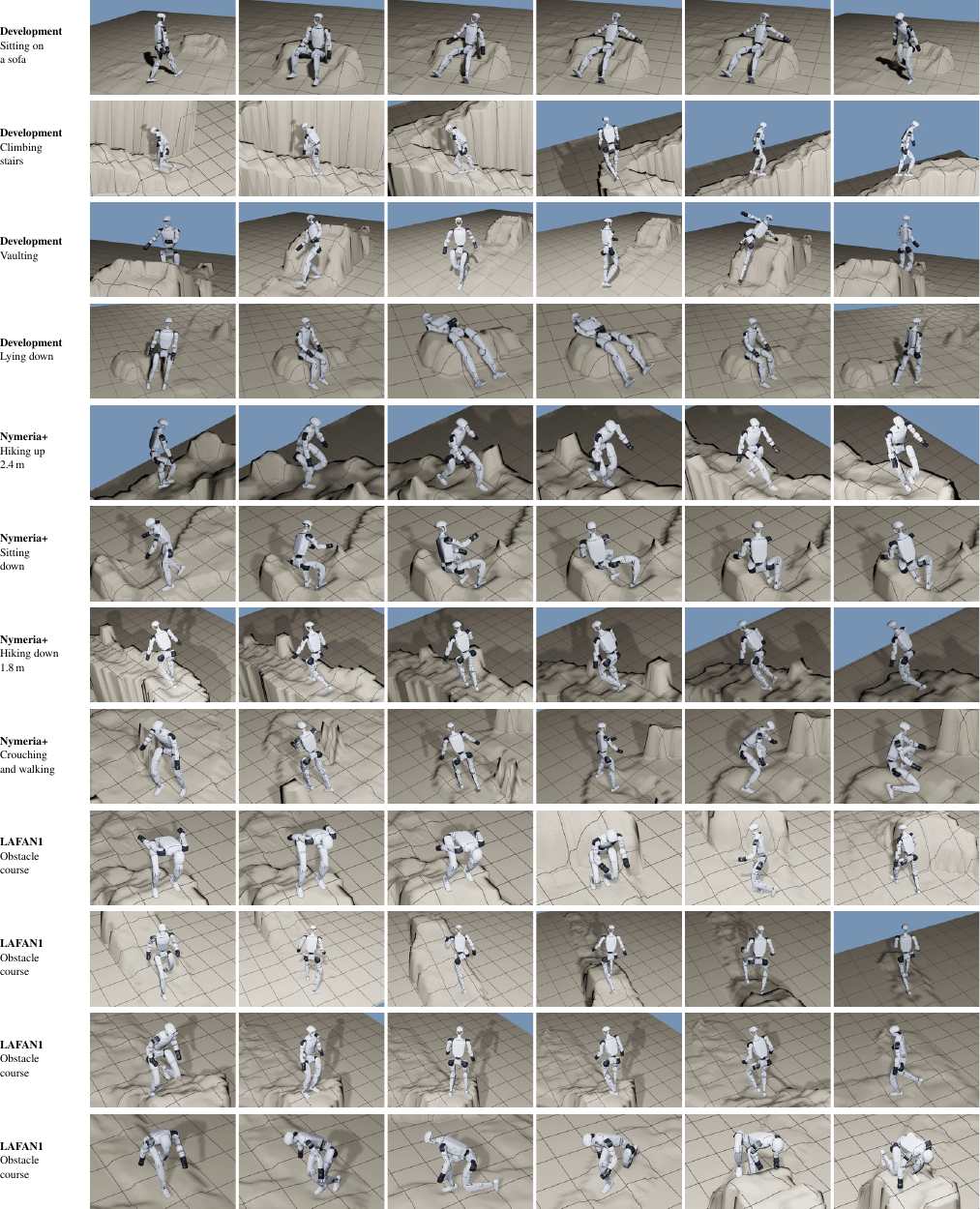}
  \caption{Six evenly spaced frames of each clip, one clip per row, cut from recorded simulator videos of
    our controllers on their generated terrain with the robot kept in view. The development clips keep
    the fixed camera of their recording, and the clips from the motion-capture datasets use a camera that
    follows the robot.}
  \label{fig:gallery-full}
\end{figure}

\paragraph{Rollout strips.}
\Cref{fig:gallery-full} shows multi-frame rollouts of our specialists on their sculpted
terrains.

\FloatBarrier

\section{Downstream Adaptation Details}
\label{app:downstream}

\subsection{Architecture and Training}
\Cref{fig:sonic-t} illustrates the architecture of \adapt{}, and \cref{tab:hyper-adapt}
lists all hyperparameters.

\paragraph{Base model.}
SONIC \citep{luo2026sonic} encodes a window of future reference frames into finite scalar
quantization tokens (FSQ, $64$ values, \citealp{mentzer2024fsq}) and decodes the tokens
together with a $10$-frame proprioceptive history (angular velocity, joint positions and
velocities, previous action, and gravity direction, $930$ values total) into $29$ joint
targets with a $9$-layer MLP. We use the released GEAR-SONIC low-latency G1
configuration, whose streaming encoder window holds $10$ frames at $0.02$\,s ($0$--$0.18$\,s
of lookahead), and freeze the encoder, the quantizer, and every decoder weight.

\paragraph{Conditioning vector.}
During training, the condition vector
$\mathbf{u}_t=[\cstar_{t:t+9},\,\boldsymbol{\delta}_t]\in[0,1]^{102}$ stacks the rollout-voted
contact labels of the ten body parts over the $10$-frame reference window with the horizontal
pelvis anchor $\boldsymbol\delta_t$, defined as the horizontal offset of the reference pelvis
from the robot's pelvis in the robot's heading frame, clipped at $0.5$\,m and perturbed by a
$0.03$\,m$/\sqrt{\mathrm{s}}$ random walk to simulate drift. With probability $0.1$ per
episode, the entire $102$-dimensional vector is set to $0.5$ so the affine mapping
$2\mathbf{u}_t-1$ maps dropped conditioning to zero.

\paragraph{Conditioned LoRA.}
Every decoder layer receives a low-rank residual
$W_\ell^{0}x+\tfrac{\alpha}{r}B_\ell A_\ell x$ with $r=16$, $\alpha=1$, and $B_\ell$
initialized to zero. In the first decoder layer, the low-rank branch reads the normalized
condition rather than the layer input,
\[
  W_1^{0}x+\frac{\alpha}{r}B_1A_1(2\mathbf{u}_t-1),
\]
following the adapter implementation of ViBe \citep{krishna2026vibe}. At initialization the
policy is identical to frozen SONIC, and the upstream FSQ token interface used by
teleoperation and vision--language--action policies remains unchanged.

\paragraph{Command-contact estimator.}
Since ground-truth labels $\cstar$ are unavailable at deployment, an online estimator
$f_\phi$ (MLP $256$--$128$, ELU) takes the exact $994$-dimensional decoder input ($64$ FSQ
token values and $930$-dimensional proprioceptive history) and predicts the $100$ contact
probabilities $\chat_{t:t+9}$ over the reference window. It is trained concurrently via
binary cross-entropy against $\cstar$ on every PPO rollout ($5$ epochs of $4$ minibatches,
Adam, learning rate $10^{-3}$). Once current-frame foot F1 ($\pm2$-frame tolerance, reached
at iteration $32$) exceeds $0.95$, the fraction of environments conditioned on $\chat$ rather
than $\cstar$ ramps linearly to $1$ over $4{,}000$ updates, reaching $0.992$ foot F1 at the
current frame and $0.991$ at the end of the $0.18$-s lookahead window. Unlike proprioceptive
estimation of current contact state \citep{ji2022concurrent}, $f_\phi$ predicts upcoming
contacts of the commanded motion over the next $0.18$\,s so the controller can prepare before
touchdown, and evaluation uses $\chat$ exclusively.

\paragraph{Critic and PPO optimization.}
A freshly initialized value critic (MLP $512$--$256$--$128$) observes SONIC's critic inputs
concatenated with the $187$-point height scan, simulated contacts $\csim$, and current
$\cstar$ ($1{,}852$ dimensions total), warming up alone for $200$ iterations. The training
reward augments SONIC's tracking objective with a confidence-weighted contact-consistency
term $\sum_k\cstar_k w_k\csim_k/\sum_k\cstar_k w_k$ while restricting SONIC's
undesired-contact penalty to body parts with $\cstar_k=0$ so sitting and kneeling are not
penalized. Both \adapt{} and the unconditioned LoRA baseline (which omits $\mathbf{u}_t$ and
trains standard $W_\ell^0 x + \frac{\alpha}{r} B_\ell A_\ell x$ residuals across all $9$
decoder layers on the same sculpted terrains) train with PPO across $4{,}096$ environments
over $64$ rotating terrain tiles ($24$-step rollouts, clip $0.2$, adaptive-KL actor learning
rate $10^{-4}$, critic learning rate $10^{-3}$), sampling segments in proportion to failure
rate mixed with $10\%$ uniform draws and minimum shares of $20\%$ flat and $10\%$ seated
clips. \Cref{tab:utility} compares both models at a matched $10{,}000$-iteration budget, and
extending \adapt{} to $16{,}500$ iterations takes $12.9$\,h on a single NVIDIA RTX~5090
($2.78$\,s per iteration).

\paragraph{Learned anchor odometry.}
To evaluate anchor tracking without simulator root positions, we also train a lightweight
proprioceptive odometer (MLP $256$--$128$ on the same $994$-dimensional token and
proprioceptive inputs, $600$ iterations with MSE loss) on rollouts of the frozen adapted
policy to predict horizontal pelvis displacement over each $20$\,ms control step ($1.7$--$1.9$\,mm
per-step error), accumulating reference-minus-predicted displacement across each evaluation
window.

\begin{table}[htbp]
  \caption{Hyperparameters and architecture settings of \adapt{}.}
  \label{tab:hyper-adapt}
  \centering
  \footnotesize
  \begin{tabularx}{\linewidth}{@{}l>{\raggedright\arraybackslash}p{4.6cm}>{\raggedright\arraybackslash}X@{}}
    \toprule
    Component & Parameter & Value \\
    \midrule
    Base       & variant, encoder, window & SONIC low-latency, G1 encoder, 10 frames at 0.02\,s (0--0.18\,s) \\
               & frozen & encoder, FSQ, all decoder weights, action std from the checkpoint \\
    \midrule
    LoRA       & layers, rank, $\alpha$, init & all 9 decoder layers, 16, 1 (scale 1/16), $B=0$ \\
               & condition & 102 values ($10\times10$ contacts $+$ 2-D anchor), as $2u-1$, first-layer delta only \\
               & condition dropout & whole condition set to 0.5 with probability 0.1 per episode \\
               & anchor & heading frame, clipped at 0.5\,m, training random walk 0.03\,m$/\sqrt{\mathrm{s}}$ \\
    \midrule
    Estimator  & input, network, output & 994 (64 token $+$ 930 proprioception), MLP 256--128 ELU, 100 logits \\
               & training & BCE vs.\ $\cstar$, Adam $10^{-3}$, 5 epochs $\times$ 4 minibatches per PPO update \\
               & transition & after foot F1 $\ge 0.95$ ($\pm2$ frames), linear ramp over 4{,}000 updates \\
    \midrule
    Critic     & network, input & MLP 512--256--128 (new), 1{,}852 (SONIC critic 1{,}645, scan 187, $\csim$ 10, $\cstar$ 10) \\
               & warm-up & 200 critic-only iterations \\
    \midrule
    Reward     & additions & contact consistency (weight 1.0), undesired contacts only where $\cstar=0$ \\
    \midrule
    PPO        & environments, rollout, epochs, minibatches & 4{,}096, 24, 5, 4 \\
               & actor lr (adaptive KL 0.01), critic lr & $10^{-4}$, $10^{-3}$ fixed \\
               & $\gamma$, $\lambda$, clip, entropy & 0.99, 0.95, 0.2, 0 \\
    \midrule
    Data       & tiles, reassignment & 64, one tile every 16 steps \\
               & sampling & by failure rate, 10\% uniform, floors flat 20\% and seat 10\% \\
               & iterations & matched budget 10{,}000 (\cref{tab:utility}), extended checkpoint 16{,}500 (2.78\,s per iteration) \\
    \midrule
    Odometer   & network, target, training & MLP 256--128 on the 994 inputs, per-step pelvis displacement, 600 iterations, MSE \\
    \bottomrule
  \end{tabularx}
\end{table}
\FloatBarrier

\subsection{Dataset Split and Evaluation Protocol}
Segments are stratified by the elevation $u_t$ of the lower foot above the segment's lowest
foothold into \emph{flat} if the $95$th percentile of $u_t$ is below $5$\,cm, \emph{terrain} if
it exceeds $0.3$\,m and $u_t>0.15$\,m on more than $30\%$ of frames, \emph{seat} if the
pelvis is labelled in contact on at least $5\%$ of frames, and \emph{mixed} otherwise.
Holding out roughly $10\%$ of recordings stratified by dominant motion type yields a
recording-disjoint split of $1{,}994$ training segments ($14.7$\,h) and $201$ held-out
segments ($2.0$\,h, \cref{tab:data}).

\begin{table}[htbp]
  \caption{Subset of \data{} used in the \adapt{} downstream study by motion stratum,
    reporting training and held-out segments and total duration.}
  \label{tab:data}
  \centering
  \footnotesize
  \begin{tabular}{lrrr}
    \toprule
    Stratum & Train & Held-out & Hours \\
    \midrule
    Flat    &     282 &  16 &  2.57 \\
    Terrain &     949 &  82 &  4.13 \\
    Seat    &      71 &  10 &  1.28 \\
    Mixed   &     692 &  93 &  8.70 \\
    \midrule
    Total   & 1{,}994 & 201 & 16.68 \\
    \bottomrule
  \end{tabular}
\end{table}

Policies are evaluated deterministically without action noise or domain randomization across
$10$-s windows initialized uniformly at random along each of the $179$ held-out segments
($1.9$\,h) lasting at least $10$\,s ($64$ terrain, $105$ flat or mixed, and $10$ seated, $8$
windows per segment, $16$ for frozen SONIC), with $95\%$ confidence intervals computed via
stratified segment-level bootstrap ($4{,}000$ resamples). A window fails if pelvis height
deviates from the reference by more than $0.25$\,m, ankle or wrist height deviates by more
than $0.25$\,m relative to posture, or simulation diverges. Frozen SONIC fails on $91.9\%$ of
held-out terrain windows ($8.1\%$ success in \cref{tab:utility}), predominantly through
pelvis-height violations ($808$ of $952$ failed windows) when a foot meets a raised step
without the body following. Extending \adapt{} from the $10{,}000$-iteration budget in
\cref{tab:utility} ($89.3\%$ overall, $92.2\%$ terrain) to $16{,}500$ iterations reaches
$90.1\%$ overall and $96.7\%$ on terrain ($74.1\%$ vs.\ $35.0\%$ on seated clips), or $87.1\%$
overall and $90.3\%$ on terrain when accumulating $\boldsymbol\delta_t$ purely from the
learned proprioceptive odometer.

\end{document}